\documentclass[sigconf]{acmart}
\usepackage{algorithm}
\usepackage{amsmath}
\usepackage{url}
\usepackage{multirow}
\usepackage{float}

\AtBeginDocument{%
  }

\setcopyright{acmlicensed}
\copyrightyear{2026}
\acmYear{2026}
\setcopyright{cc}
\setcctype{by}
\acmConference[KDD'26] {Proceedings of the 32nd ACM SIGKDD Conference on Knowledge Discovery and Data Mining V.1}{August 9--13, 2026}{Jeju Island, Republic of Korea}
\acmBooktitle{Proceedings of the 32nd ACM SIGKDD Conference on Knowledge Discovery and Data Mining V.1 (KDD'26), August 9--13, 2026, Jeju Island, Republic of Korea}
\acmPrice{}
\acmDOI{10.1145/3770854.3780194}
\acmISBN{979-8-4007-2258-5/2026/08}

\begin{document}

\title{How Do Graph Signals Affect Recommendation: Unveiling the Mystery of Low and High-Frequency Graph Signals}

\author{Feng Liu}
\authornote{Feng Liu and Hao Cang contributed equally to this work.}
\affiliation{%
  \department{School of Computer Science and Technology}
  \institution{Soochow University}
  \city{Suzhou}
  \country{China}
}
\email{fliuliufeng@stu.suda.edu.cn}

\author{Hao Cang}
\authornotemark[1]
\affiliation{%
  \department{School of Computer Science and Technology}
  \institution{Soochow University}
  \city{Suzhou}
  \country{China}
  }
\email{hcangada@stu.suda.edu.cn}

\author{Huanhuan Yuan}
\affiliation{%
  \department{School of Computer Science and Technology}
  \institution{Soochow University}
  \city{Suzhou}
  \country{China}
  }
\email{hhyuan@stu.suda.edu.cn}

\author{Jiaqing Fan}
\affiliation{%
  \department{School of Computer Science and Technology}
  \institution{Soochow University}
  \city{Suzhou}
  \country{China}
  }
\email{jqfan@suda.edu.cn}

\author{Yongjing Hao}
\affiliation{%
  \department{School of Electronic and Information Engineering}
  \institution{Suzhou University of Science and Technology}
  \city{Suzhou}
  \country{China}
}
\email{yjhao@mail.usts.edu.cn}

\author{Fuzhen Zhuang}
\affiliation{%
  \department[0]{Institute of Artificial Intelligence, Beihang University}
  \department[1]{State Key Laboratory of Complex \& Critical Software Environment}
  \institution{Beihang University}
  \city{Beijing}
  \country{China}
}
\email{zhuangfuzhen@buaa.edu.cn}

\author{Guanfeng Liu}
\affiliation{%
 \department{School of Computing}
 \institution{Macquarie University}
 \city{Sydney}
 \country{Australia}}
 \email{guanfeng.liu@mq.edu.au}

\author{Pengpeng Zhao}
\authornote{Corresponding author.}
\affiliation{%
  \department{School of Computer Science and Technology}
  \institution{Soochow University}
  \city{Suzhou}
  \country{China}}
\email{ppzhao@suda.edu.cn}

\renewcommand{\shortauthors}{Feng Liu et al.}

\begin{abstract}
  Spectral graph neural networks (GNNs) are highly effective in modeling graph signals, with their success in recommendation often attributed to low-pass filtering. However, recent studies highlight the importance of high-frequency signals. The role of low-frequency and high-frequency graph signals in recommendation remains unclear.
  This paper aims to bridge this gap by investigating the influence of graph signals on recommendation performance. We theoretically prove that the effects of low-frequency and high-frequency graph signals are equivalent in recommendation tasks, as both contribute by smoothing the similarities between user-item pairs. To leverage this insight, we propose a frequency signal scaler, a plug-and-play module that adjusts the graph signal filter function to fine-tune the smoothness between user-item pairs, making it compatible with any GNN model.
  Additionally, we identify and prove that graph embedding-based methods cannot fully capture the characteristics of graph signals. To address this limitation, a space flip method is introduced to restore the expressive power of graph embeddings. Remarkably, we demonstrate that either low-frequency or high-frequency graph signals alone are sufficient for effective recommendations. Extensive experiments on four public datasets validate the effectiveness of our proposed methods.
  Code is avaliable at \url{https://github.com/mojosey/SimGCF}.
\end{abstract}




\maketitle

\section{Introduction}
Graph neural networks (GNNs) have been widely used in collaborative filtering-based recommender systems \cite{hao2024meta,he2017neural,he2020lightgcn} due to the ability to effectively capture high-level information on graphs. Early GNN-based recommendation primarily rely on spatial domain GNNs \cite{he2017neural,wang2020disentangled,liu2021interest}. They explore the construction of graph structures, the design of message propagation mechanisms and aggregation strategies from a spatial perspective, achieving remarkable results.
\begin{figure}
\centering
\begin{minipage}{0.23\textwidth}
  \centering
  \includegraphics[width=1\linewidth]{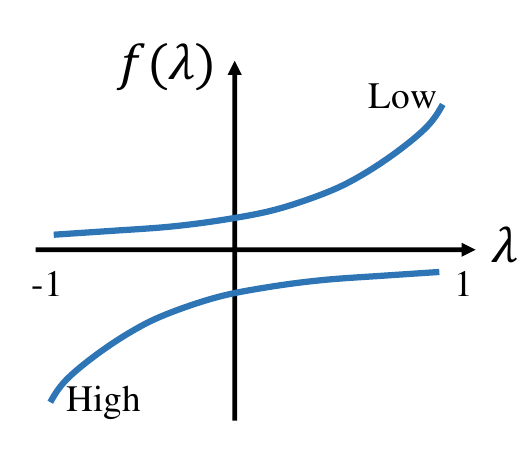}
  \centerline{(a)}
\end{minipage}%
\centering
\begin{minipage}{0.27\textwidth}
  \centering
  \includegraphics[width=1\linewidth]{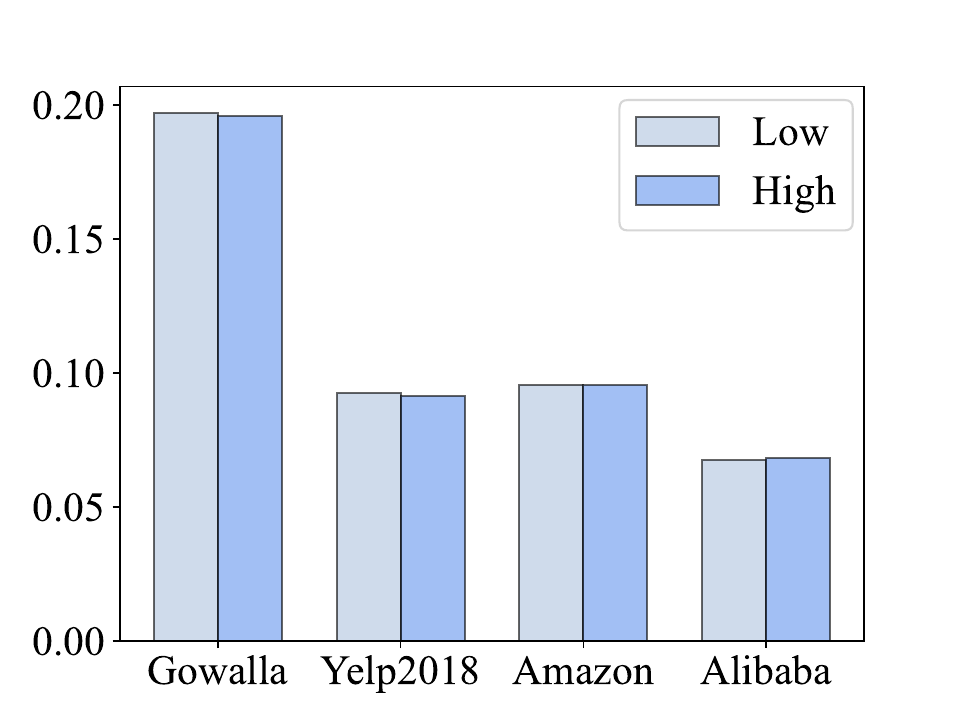}
  \centerline{(b)}
\end{minipage}
\caption{Analyze the impact of low-frequency and high-frequency graph signals on recommendation. (a) The waveforms of low-pass function in quadrant I and high-pass function in quadrant III. (b) The performance of low-pass and high-pass GNNs on four datasets, evaluated on Recall@20.}
\label{fig:high_low_recall}
\Description{ }
\end{figure}

Recently, recommendations based on spectral domain GNNs have garnered significant attention \cite{luo2024spectral,peng2024powerful}, focusing on exploring the role of graph signals in recommendation. Researchs mainly include matrix decomposition-based methods \cite{shen2021powerful,liu2023personalized,xia2025frequency} and graph embedding-based methods \cite{wang2019neural,he2020lightgcn,guo2023manipulating}. The former is computationally efficient but difficult to scale, while the latter approximates graph filters through polynomials and is more suitable for large-scale recommendations. Studies have shown that most current GNN-based recommendation models function as low-pass filters \cite{he2017neural,he2020lightgcn}. These studies indicate that low-frequency graph signals play a key role in GNN-based recommendation models \cite{shen2021powerful,qin2024polycf}, as they can smooth the similarities between the embeddings of connected nodes. However, a study suggests that high-frequency graph signals also contribute to recommendation \cite{guo2023manipulating}. They find that low-frequency and high-frequency graph signals in the training data are strongly linearly correlated with those in the test data, and they propose a graph embedding method based on Jacobi polynomial bases to leverage graph signals of different frequencies. 

However, \textbf{how do low-frequency and high-frequency graph signals affect recommendation} remains an open question. Limited by the exploration of graph signals across different frequencies in recommender systems, the underlying mechanisms of graph signals in recommendation remain largely unexplored.

In this paper, we explore how graph signals affect recommendation. First, we explore the impact of low and high-frequency signals in recommendation. As shown in Figure \ref{fig:high_low_recall}(a), we utilize the low-frequency graph signal \textbf{Low} and the high-frequency graph signal \textbf{High} with the same absolute value of waveform for recommendation. As shown in Figure \ref{fig:high_low_recall}(b), their performances are very close, indicating that their impact on recommendations is equivalent. We theoretically prove that low and high-frequency graph signals with the same absolute value of waveform have an equal impact on recommendation. Graph signals affect recommendation by changing the similarity between user-item pairs, the key lies in the waveform of the graph signal filter function. Based on the above experimental and theoretical findings, we introduce a frequency signal scaler constructed with a learnable monomial basis to further refine the waveform of the graph signal filter function in the original GNN model.
As a plug-and-play plugin, it can be applied to any GNN to enhance the performance of downstream tasks.

In addiation, we theoretically demonstrate that existing spectral GNNs cannot fully express the characteristics of graph signals. This limitation arises from the fact that current spectral GNNs represent graph signals through graph embeddings \cite{liu2024simgcl,li2023homogcl,yang2023generative,wang2022adagcl}. 
We find that graph signals exhibit four distinct types of features and the graph embedding signals can only capture half of these features, with the other features being hidden. Therefore, we proposes a space flip method to restore the expressiveness of graph embedding signals by flipping the original graph embeddings.  

The contributions of this paper are as follows:

\begin{itemize}
\item We study the impact of graph signals on recommendation and prove that low-frequency and high-frequency graph signals are equivalent in recommendation, both influencing recommendation performance by altering the similarity between user-item pairs.

\item We analyze the differences between graph embedding signals and graph signals, finding that current graph embedding signals cannot fully capture the characteristics of graph signals, and propose a space flip method to enhance their expressive power.

\item Extensive experiments on four widely used benchmark datasets demonstrate that our approach effectively improves upon state-of-the-art models.
\end{itemize}

\section{Related Works}
\label{novety_and_compare_with_related_works}
\subsection{Spectral Graph Collaborative Filtering based on Matrix Decomposition}
Spectral graph collaborative filtering based on matrix decomposition primarily investigates the properties of graph signals on user-item graphs derived through singular value decomposition (SVD) or eigenvalue decomposition. GC\_CF \cite{shen2021powerful} demonstrates the critical role of low-frequency graph signals in CF and proposes an algorithm that combines linear and ideal low-pass filters to directly filter the interaction matrix. PGSP \cite{liu2023personalized} argues that high-frequency graph signals also play an important role in representing user preferences and introduces a personalized graph signal processing method to incorporate high-frequency signals. SGFCF \cite{xia2025frequency} studies the expressive power of graph signal filters used in current decomposition-based spectral collaborative filtering methods and points out that linear filters cannot fit arbitrary embeddings. Although matrix decomposition-based spectral methods enable simple and efficient filtering of graph signals, they require eigenvalue decomposition, which limits their efficiency in large-scale recommendation. Moreover, they fail to encode structural information into vectors, which restricts their scalability to downstream tasks.

\subsection{Spectral Graph Collaborative Filtering based on Graph Embedding}
Spectral graph collaborative filtering based on graph embedding approximates graph signal filters by using different polynomial bases. In theory, any graph signal filter can be approximated by a high-order polynomial basis. Early graph embedding-based collaborative filtering models are low-pass graph signal filters. For example, NGCF \cite{wang2019neural} and LightGCN \cite{he2020lightgcn} are both low-pass filtering graph models based on monomial bases. JGCF \cite{guo2023manipulating} points out that high-frequency graph signals are also crucial in recommendation, and proposes a hybrid filtering GCF model based on Jacobi polynomial bases. Current graph embedding methods \cite{chen2023bridging,he2022convolutional,chen2024polygcl} generally assume that the graph signals represented by embeddings are equivalent to the graph signals, without thoroughly exploring the differences between them. However, graph embedding can only capture part of the graph signal characteristics. Moreover, the impact of graph signals at different frequencies within embeddings on recommendation performance remains an open question.

\section{Preliminary}
\subsection{Spectral Graph Neural Network}
We denote the user set as $\mathcal{U}$, the item set as $\mathcal{I}$, and the nodes set as $N=|\mathcal{U}|+|\mathcal{I}|$. The interaction matrix of users and items is $\mathbf{R}\in\mathbb{R}^{|\mathcal{U}|\times |\mathcal{I}|}$ where $\mathbf{R}_{ij}=1$ if the i-th user and the j-th item have an interaction. The adjacency matrix of the graph in the graph-based recommendation can be expressed as follows:
\begin{equation}
\mathbf{A}= \left[\begin{array}{cc} 
\mathbf{0}&\mathbf{R}\\
\mathbf{R}^{T}&\mathbf{0}
\end{array}\right] 
\end{equation}

Spectral GNN performs graph convolution in the Laplacian spectral domain. The Laplacian matrix can be defined as $\mathbf{L}=\mathbf{D}-\mathbf{A}$, where $\mathbf{D}\in\mathbb{R}^{N\times N}$ represents the degree matrix of the graph. Its normalized form can be expressed as $\hat{\mathbf{L}}=\mathbf{I}-\hat{\mathbf{A}}$, where $\hat{\mathbf{A}} =\mathbf{D}^{-\frac{1}{2}}\mathbf{A}\mathbf{D}^{-\frac{1}{2}}$. As $\hat{\mathbf{L}}$ is a symmetric positive definite matrix, we can obtain its eigenvalues and corresponding eigenvectors through eigenvalue decomposition, $\hat{\mathbf{L}}=\mathbf{U}\mathbf{\Lambda}\mathbf{U}^{T}$, where $\mathbf{U}\in \mathbb{R}^{N\times N}$ is the eigenvector and $\mathbf{\Lambda}\in\mathbb{R}^{N\times N }$ is the corresponding eigenvalue, the eigenvalue range of the Laplacian matrix is [0,2]. The graph Fourier transform of the signal $\mathbf{x}\in\mathbb{R}^{N}$ can be defined as $\mathbf{\hat{x}}=\mathbf{U}^{T}\mathbf{x}$ and the inverse transform is $\mathbf{x}= \mathbf{U}\hat{\mathbf{x}}$. This transformation enables the formulation of operations such as filtering in the spectral domain. The filtering operation on signal $\mathbf{x}$ can be defined as:
\begin{equation}
    \mathbf{y} = f(\hat{\mathbf{L}})\mathbf{x}=\mathbf{U}f(\mathbf{\mathbf{\Lambda}})\mathbf{U}^T\mathbf{x}
\end{equation}

Directly computing $f(\mathbf{\Lambda})$ is difficult due to the eigenvalue decomposition of large-scale matrices is very time-consuming. Therefore, spectral GNN often uses some polynomials to approximate $f(\mathbf{\Lambda})$:
\begin{equation}
    f(\mathbf{\Lambda})=\sum_{i=0}^{n}\theta_{i}\mathbf{P}_{i}(\mathbf{\Lambda})
\end{equation}
where $\theta_{i}$ is usually a learnable or fixed scalar, $\mathbf{P}_{i}(\mathbf{\Lambda})$ usually uses various forms of polynomial bases, such as Monomials, Chebyshev polynomials and Jacobi polynomial bases.
\begin{figure*}[ht]
    \centering
    \includegraphics[width=1\linewidth]{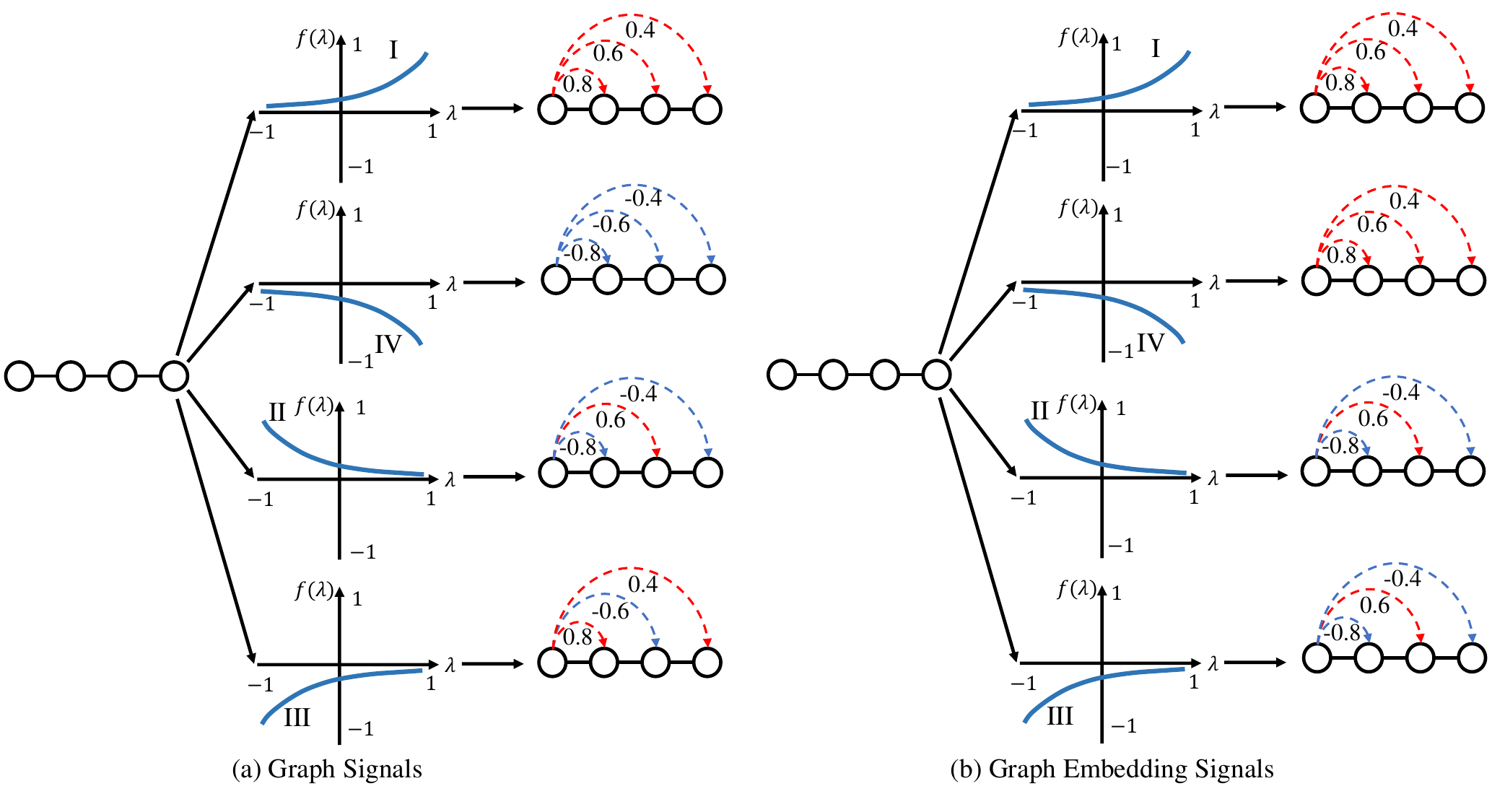}
    \caption{The Characteristics of graph signals and graph embedding signals.}
    \label{fig:gs}
    \Description{ }
\end{figure*}

\subsection{Graph Signal Filter in Recommendation}

Different from spectral graph neural networks, GNN-based recommendation typically perform graph signal filtering on the normalized adjacency matrix $\hat{\mathbf{A}}$, influenced by the removal of self-loops as proposed in LightGCN \cite{he2020lightgcn}. Consequently, the eigenvalues of $\hat{\mathbf{A}}$ fall within the range of [-1, 1].

\textbf{Definition 1} \emph{Graph Signal} (GS). The graph signal refers to a series of eigenvalues $\mathbf{\Lambda}$ ranging from [-1,1] obtained by eigenvalue decomposition of the adjacency matrix $\hat{\mathbf{A}}$. It exhibits high-frequency characteristics when the eigenvalue is close to -1 and low-frequency characteristics when it is close to 1. Usually, the spectral operation on the graph signals can be performed by the graph signals filtering function $f(\lambda)$:
\begin{equation}
\mathbf{S}_{1}=f(\hat{\mathbf{A}})=\mathbf{U}f(\mathbf{\Lambda})\mathbf{U}^{T}=\sum_{i=0}^{n}\alpha_{i}\mathbf{U}\mathbf{\Lambda}^{i}\mathbf{U}^{T}=\sum_{i=0}^{n}\alpha_i\hat{\mathbf{A}}^{i}
\end{equation}

\textbf{Definition 2} \emph{Graph Embedding Signal} (GES). The graph embedding signal refers to compressing the features of the graph signal into vectors by embedding methods, $\mathbf{E}=f(\hat{\mathbf{A}})\mathbf{E}^{0}$. The features of graph signals are approximately restored by calculating the similarity between vectors:
\begin{equation}
\mathbf{S}_{2}=\mathbf{E}\mathbf{E}^{T}=(f(\hat{\mathbf{A}})\mathbf{E}^{0})(f(\hat{\mathbf{A}})\mathbf{E}^{0})^{T}
\end{equation}

\textbf{Definition 3} \emph{Graph signal filtering function}. The graph signal filter function $f(\lambda)$ refers to the transformation of the original graph signal $\lambda\in \Lambda$ to obtain a new graph signal, usually approximated by a polynomial bases. Given a set of polynomial coefficients $\alpha_i>0$, where $i\in[0,n]$, we can obtain four different quadrants of graph signal filtering function, as shown in Figure \ref{fig:gs}:
\begin{itemize}
    \item \textbf{Low(I)}, Low frequency filter function in quadrant I: $f(\lambda)=\alpha_0\lambda^0+\alpha_1\lambda^1+...+\alpha_n\lambda^n=\sum_{i=0}^{n}\alpha_i\lambda^i$.
    \item \textbf{Low(IV)}, Low frequency filter function in quadrant IV: $f(\lambda)=-\alpha_0\lambda^0-\alpha_1\lambda^1-...-\alpha_n\lambda^n=\sum_{i=0}^{n}-\alpha_i\lambda^i$.
    \item \textbf{High(II)}, High frequency filter function in quadrant II: $f(\lambda)=\alpha_0\lambda^0-\alpha_1\lambda^1+...+(-1)^{n}\alpha_n\lambda^n=\sum_{i=0}^{n}(-1)^{i}\alpha_i\lambda^i$.
    \item \textbf{High(III)}, High frequency filter function in quadrant III: $f(\lambda)=-\alpha_0\lambda^0+\alpha_1\lambda^1+...+(-1)^{n+1}\alpha_n\lambda^n=\sum_{i=0}^{n}(-1)^{i+1}\alpha_i\lambda^i$.
\end{itemize}

\section{How do Graph Signals Work in Recommender Systems?}
This section aims to investigate how graph signals function affect recommender systems. Specifically, we first explore the ability of graph embedding signals methods to express graph signals. Then we analyze the key factors that affect recommendations with graph signals of different frequencies. These analyses collectively provide a theoretical basis for understanding the role of graph signal frequency in recommendation performance.

\subsection{Is Graph Signal Equal to Graph Embedding Signal?}
Collaborative Filtering (CF) assumes that users are similar to the items they have interacted with, and is usually viewed as a link prediction task on a homogeneous graph. Study has shown that low-frequency Graph Embedding Signals (GES) play a key role in this task \cite{shen2021powerful}. Recently, a study shows that there is a strong linear correlation between the high-frequency Graph Signals (GS) of the interaction graphs in the training set and the test set. They proposed a GES method based on the Jacobi polynomial bases to utilize the high-frequency GS \cite{guo2023manipulating}. This seems to contradict the assumption of CF. As high-frequency GES cause connected users and items to be dissimilar, while CF assumes that there is similarity between interacting users and items. As shown in Table \ref{tab:high-and-low}, the experimental results indicate that high-frequency GES have a much smaller impact on the performance compared to low-frequency GES. 
In fact, we find that there is a huge difference between the GES and the GS, and proved that the GES can only express part of the characteristics of the GS. The high-frequency GS that are beneficial for recommender systems cannot be expressed by GES.
\begin{table}[ht]
    \centering
    \caption{The impact of high-frequency graph embedding signals and low-frequency graph embedding signals.}
    \label{tab:high-and-low}
    \resizebox{1\linewidth}{!}{
    \begin{tabular}{c|cc|cc}
    \toprule
        \multirow{2}*{Datasets}& \multicolumn{2}{c|}{High} & \multicolumn{2}{c}{Low}\\
        & Recall@10 & NDCG@10 & Recall@10 & NDCG@10\\
        \midrule
        \textbf{Gowalla} & 0.0217 & 0.0134 & 0.1362 & 0.0987 \\ 
        \textbf{Yelp2018} & 0.0097 & 0.0071 & 0.0562 & 0.0453 \\ 
        \textbf{Amazon-Books} & 0.0107 & 0.0087 & 0.0620 & 0.0508 \\
        \textbf{Alibaba-iFasion} & 0.0129 & 0.0063 & 0.0450 & 0.0244\\
        \bottomrule
    \end{tabular}
    }
\end{table}

First, we describe the specific characteristics of graph signals and graph embedding signals.

\textbf{Theorem 1} \emph{Characteristics of low-frequency GS.} The low-frequency GS $\mathbf{S}_{low}^{1}$ exhibits a stable changing characteristic, and the absolute value of the similarity between nodes decays with the increase of the path length. When $f(\lambda)>0$, all connected nodes will become similar, and when $f(\lambda)<0$, all connected nodes will become dissimilar. As shown in Figure \ref{fig:gs}(a), the signal filtering functions $f(\lambda)$ in quadrants I and IV.

\textbf{Theorem 2} \emph{Characteristics of high-frequency GS.} The high-frequency GS $\mathbf{S}_{high}^{1}$ exhibits the characteristics of frequent changes, and the absolute value of the similarity between nodes decays with the increase of the path length. When $f(\lambda)>0$, the odd-order nodes will be dissimilar and the even-order nodes will be similar. When $f(\lambda)<0$, the odd-order nodes will be similar and the even-order nodes will be dissimilar. As shown in Figure \ref{fig:gs}(a), the signal filtering functions $f(\lambda)$ in quadrants II and III.

\textbf{Theorem 3} \emph{Characteristics of low-frequency GES.} The low-frequency GES $\mathbf{S}_{low}^{2}$ exhibits a stable changing characteristic, and the absolute value of the similarity between nodes decays with the increase of the path length. When $f(\lambda)>0$ or $f(\lambda)<0$, it makes all connected nodes similar. As shown in Figure \ref{fig:gs}(b), the signal filtering functions $f(\lambda)$ in quadrants I and IV.

\textbf{Theorem 4} \emph{Characteristics of high-frequency GES.} The high-frequency GES $\mathbf{S}_{high}^{2}$ exhibits the characteristics of frequent changes, and the absolute value of the similarity between nodes decays with the increase of the path length. When $f(\lambda)>0$ or $f(\lambda)<0$, it will make the odd-order nodes dissimilar and the even-order nodes similar. As shown in Figure \ref{fig:gs}(b), the signal filter function $f(\lambda)$ in quadrants II and III.

From the above theorems, we can draw the following theoretical corollary.


\textbf{Corollary 1} \emph{The GES can only preserve the characteristics of the GS with $f(\lambda)>0$, while the GES will hide the characteristics of the GS with $f(\lambda)<0$.}

\textbf{Proof of Corollary 1} Given a graph signal filter function with $f(\lambda)>0$ and $f_a(\hat{\mathbf{A}})=\sum_{i=0}^{n}\alpha_i\hat{\mathbf{A}}^{i}$, the corresponding graph embedding signal is $\mathbf{S}_{a}^{2}=\mathbf{E}_{a}\mathbf{E}_{a}^{T}=(f_a(\hat{\mathbf{A}})\mathbf{E}^0)(f_a(\hat{\mathbf{A}})\mathbf{E}^0)^{T}$. Similarly, the corresponding graph signal filtering function for $f(\lambda)<0$ is $f_b(\hat{\mathbf{A}})=-f_a(\hat{\mathbf{A}})=\sum_{i=0}^{n}-\alpha_i\hat{\mathbf{A}}^{i}$, and the graph embedding signal is $\mathbf{S}_{b}^2=\mathbf{E}_b\mathbf{E}_b^{T}=(f_b(\hat{\mathbf{A}})\mathbf{E}^0)(f_b(\hat{\mathbf{A}})\mathbf{E}^0)^{T}$. Based on the above conditions, we can derive the following derivation:
\begin{equation}
\begin{aligned}
\mathbf{S}_{a}^{2}
&=\mathbf{E}_{a}\mathbf{E}_{a}^{T}=(f_{a}(\mathbf{\hat{A}})\mathbf{E}^0)(f_a(\mathbf{\hat{A}})\mathbf{E}^0)^T \\
&=(\alpha_0\mathbf{\hat{A}}^{0}\mathbf{E}^0+...+\alpha_n\mathbf{\hat{A}}^n\mathbf{E}^0)(\alpha_0\mathbf{\hat{A}}^{0}\mathbf{E}^0+...+\alpha_n\mathbf{\hat{A}}^n\mathbf{E}^0)^{T} \\
&=(-\alpha_0\mathbf{\hat{A}}^{0}\mathbf{E}^0-...-\alpha_n\mathbf{\hat{A}}^n\mathbf{E}^0)(-\alpha_0\mathbf{\hat{A}}^{0}\mathbf{E}^0-...-\alpha_n\mathbf{\hat{A}}^n\mathbf{E}^0)^{T} \\
&=(-f_a(\mathbf{\hat{A}})\mathbf{E}^0)(-f_a(\mathbf{\hat{A}})\mathbf{E}^0)^T\\
&=(f_b(\mathbf{\hat{A}})\mathbf{E}^0)(f_b(\mathbf{\hat{A}})\mathbf{E}^0)^T=\mathbf{E}_{b}\mathbf{E}_b^T=\mathbf{S}_b^2
\end{aligned}
\end{equation}

\label{explain_for_space_flip}
The above derivation shows that GES fails to capture the negative sign when representing graph signals with $f(\lambda)<0$. We assume $\mathbf{X}=f_a(\mathbf{\hat{A}})\mathbf{E}^0$, $-\mathbf{X}=-f_a(\mathbf{\hat{A}})\mathbf{E}^0=f_b(\mathbf{\hat{A}})\mathbf{E}^0$. In this case, $\mathbf{S}_{a}^{2}=\mathbf{X}^2$ and $\mathbf{S}_{b}^{2}=(-\mathbf{X})^2$. However, the GES corresponding to $f(\lambda)<0$ should actually be $-\mathbf{X}^2$. Therefore, a negative sign needs to be applied to $\mathbf{S}_b^2$ to recover the correct GES. See the Appendix \ref{lab:gs_ges} for details.

\subsubsection{Space Flip}
Through the proof of \textbf{Corollary 1}, we can easily observe that GES fails to express negative sign when calculating the similarity matrix between nodes, which suppresses some of the properties of GS. Therefore, we need to add a negative sign in front when calculating these features to restore the original properties:
\begin{equation}
\mathbf{S}_{2}=-\mathbf{E}\mathbf{E}^{T}=(-\mathbf{E})\mathbf{E}^{T}=\mathbf{E}_{f}\mathbf{E}^{T}
\end{equation}

This operation can be viewed as a space flip of the original graph embedding $\mathbf{E}$ to obtain a new embedding $\mathbf{E}_{f}=-\mathbf{E}$, and then calculating its similarity with the original graph embedding $\mathbf{E}$. By space flip, we can recover the hidden high-frequency features of GS in quadrant III. As shown in Table \ref{tab:recover}, we can find that compared with the features of high-frequency GS in quadrant II, the features of high-frequency GS in quadrant III have a significant impact on the recommendation performance, which is at the same level as the impact of low-frequency GS in quadrant I.
\begin{table}[ht]
    \centering
    \caption{The performance of GS under four different quadrants on Recall@10 after space flip.}
    \label{tab:recover}
    \resizebox{1\linewidth}{!}{
    \begin{tabular}{c|c|c|c|c}
    \toprule
        Datasets & low(IV) & low(I) & high(II) & high(III) \\ 
        \midrule
        \textbf{Gowalla} & 0.0219 & 0.1362 & 0.0217 & 0.1362 \\
        \textbf{Yelp2018} & 0.0113 & 0.0562 & 0.0097 & 0.0561 \\
        \textbf{Amazon-Books} & 0.0102 & 0.0620 & 0.0107 & 0.0619 \\
        \textbf{Alibaba-iFasion} & 0.0130 & 0.0450 & 0.0129 & 0.0453\\ 
        \bottomrule
    \end{tabular}
    }
\end{table}

\subsection{How do Low-frequency and High-frequency Graph Signals Affect Recommendation?}
From the above analysis, we can recover the High(III) GS by GES, which is beneficial for recommendation. However, we identify an issue: High(III) and Low(I) with the same absolute value exhibit identical performance in recommendation, as shown in Table \ref{tab:recover}. This prompts us to consider how do graph signals of different frequencies affect recommendation. In fact, we find that low-frequency GS and high-frequency GS with the same absolute value have an equivalent impact on recommendation. Both influence the recommendation by affecting the similarity between user-item pairs, with the key lying in the waveform of the filter function.

\textbf{Corollary 2} \emph{The low-frequency graph signal Low(I) and the high-frequency graph signal High(III) with the same absolute value are equivalent in recommendation, and Low(IV) and High(II) are also equivalent.}

\textbf{Proof of Corollary 2} As given in \textbf{Definition 3}, the filter function of Low(I), $f_{\mathbf{I}}(\lambda)=\sum_{i=0}^{n}\alpha_i\lambda^i$, and the filter function of High(III), $f_{\mathbf{III}}(\lambda)=\sum_{i=0}^{n}(-1)^{i+1}\alpha_i\lambda^i$. We perform graph embedding on it $\mathbf{E}_{\mathbf{I}}=f_{\mathbf{I}}(\lambda)\mathbf{E}^0$, $\mathbf{E}_{\mathbf{III}}=f_{\mathbf{III}}(\lambda)\mathbf{E}^0$. Then calculate the graph embedding signal. When calculating the graph embedding signal of High(III), we perform space flip on it. We can get the graph embedding signal of Low(I), $\mathbf{S}_{\mathbf{I}}=\mathbf{E}_{\mathbf{I}}\mathbf{E}_{\mathbf{I}}^{T}$, and the graph embedding signal of High(III), $\mathbf{S}_{\mathbf{III}}=-\mathbf{E}_{\mathbf{III}}\mathbf{E}_{\mathbf{III}}^{T}$. We expand $\mathbf{S}_{\mathbf{I}}$ and $\mathbf{S}_{\mathbf{III}}$ to get the following expression:
\begin{equation}
\mathbf{S}_{\mathbf{I}}=
\left(
\begin{aligned}
(1)^{0}\alpha_0\alpha_0\mathbf{E}^0(\mathbf{E}^0)^T+&...+(1)^{n}\alpha_0\alpha_n\mathbf{E}^0(\mathbf{E}^n)^T\\
+&...+\\
(1)^{n}\alpha_n\alpha_0\mathbf{E}^n(\mathbf{E}^0)^T+&...+(1)^{2n}\alpha_n\alpha_n\mathbf{E}^n(\mathbf{E}^n)^T
\end{aligned}
\right)
\end{equation}
\begin{equation}
\mathbf{S}_{\mathbf{III}}=
\left(
\begin{aligned}
(-1)^{3}\alpha_0\alpha_0\mathbf{E}^0(\mathbf{E}^0)^T+&...+(-1)^{n+3}\alpha_0\alpha_n\mathbf{E}^0(\mathbf{E}^n)^T\\
+&...+\\
(-1)^{n+3}\alpha_n\alpha_0\mathbf{E}^n(\mathbf{E}^0)^T+&...+(-1)^{2n+3}\alpha_n\alpha_n\mathbf{E}^n(\mathbf{E}^n)^T
\end{aligned}
\right)
\end{equation}
where $\mathbf{E}^n=\mathbf{\hat{A}}^{n}\mathbf{E}^0$. By comparison, we can find that the odd-order polynomial terms of $\mathbf{S}_{\mathbf{I}}$ and $\mathbf{S}_{\mathbf{III}}$ are the same, $\mathbf{S}_{ij}^{\mathbf{odd(I)}}=\alpha_i\alpha_j\mathbf{E}^i\mathbf{E}^j$, $\mathbf{S}_{ij}^{\mathbf{odd(III)}}=\alpha_i\alpha_j\mathbf{E}^i\mathbf{E}^j$, when the sum of the subscripts is odd. When the sum of the subscripts is even, their polynomials are opposite, $\mathbf{S}_{ij}^{\mathbf{even(I)}}=\alpha_i\alpha_j\mathbf{E}^i\mathbf{E}^j$, $\mathbf{S}_{ij}^{\mathbf{even(III)}}=-\alpha_i\alpha_j\mathbf{E}^i\mathbf{E}^j$. We use odd-order polynomials and even-order polynomials of Low(I) and High(III) with the same absolute value for recommendation respectively.
\begin{table}[!ht]
    \centering
    \caption{The performance of the sum of odd-order and even-order polynomials of Low(I) and High(III) with the same absolute value on Recall@10.}
    \label{tab:odd_even}
    \resizebox{1\linewidth}{!}{
    \begin{tabular}{c|c|c|c|c}
    \toprule
        Datasets & even(I) & odd(I) & even(III) & odd(III)\\
        \midrule
        \textbf{Gowalla} & 0.0754 & 0.1224 & 0.0754 & 0.1234 \\
        \textbf{Yelp2018} & 0.0154 & 0.0451 & 0.0154 & 0.0447 \\
        \textbf{Amazon-Books} & 0.0291 & 0.0438 & 0.0296 & 0.0448 \\
        \textbf{Alibaba-iFasion} & 0.0330 & 0.0454 & 0.0334 & 0.0461 \\
        \bottomrule
    \end{tabular}
    }
\end{table}

As shown in Table \ref{tab:odd_even}, we can find that the performance of odd-order odd(I) and odd(III) is very close, which is due to $\mathbf{S}_{ij}^{\mathbf{odd(I)}}=\mathbf{S}_{ij}^{\mathbf{odd(III)}}$. The even-order even(I) is opposite to even(III), $\mathbf{S}_{ij}^{\mathbf{even(I)}}=-\mathbf{S}_{ij}^{\mathbf{even(III)}}$, but the performance in recommendation is still very close. In fact, this is because the even-order polynomial $\mathbf{E}^i(\mathbf{E}^j)^T$ emphasizes the similarity between even-order node pairs on the graph, such as user-user and item-item pairs in recommendations, while the loss function in recommendation systems aims to minimize the similarity between user-item pairs. They are orthogonal and do not conflict, which leads to the equivalence of $\mathbf{S}_{ij}^{\mathbf{even(I)}}$ and $\mathbf{S}_{ij}^{\mathbf {even(III)}}$ in the training process of the recommendation model. Therefore, Low(I) and High(III) with the same absolute value are equivalent in recommendation, the same principle applies to Low(IV) and High(II). For details, please refer to the Appendix \ref{lab:gs_in_rec}.

\subsubsection{Frequency Signal Scaler}
\label{lab:FSS}
From Table \ref{tab:odd_even}, we can see that the performance of $\mathbf{odd(I)}$ and $\mathbf{odd(III)}$ is much higher than that of $\mathbf{even(I)}$ and $\mathbf{even(III)}$. This is due to the fact that the odd-order polynomial $\mathbf{E}^i(\mathbf{E}^j)^T$ has an important impact on the similarities between user-item pairs in recommendation, which directly affects the performance of recommendation. In fact, the waveform of the GNN's graph signal filter function $f(\lambda)$ will directly affect the weights of the odd-order polynomial coefficients. However, the current GNN-based recommendation model is still not flexible enough to adjust the waveform of $f(\lambda)$. Therefore, we propose a frequency signal scaler that can flexibly adjust the waveform of the original GNN filter function. As a plug-and-play plug-in, it can be integrated into other GNN models.

Inspired by the sigmoid function waveform, we propose a frequency signal scaling function $g(\lambda)$ to extract the frequency signal that needs to be adjusted and then scale it:
\begin{equation}
g(\lambda)=\frac{\mu}{1+e^{\alpha(\lambda+\beta)}}
\end{equation}
where $\alpha\in R$, $|\alpha|$ are used to control the steepness of the waveform. When $\alpha<0$, it is used to extract low-frequency signals. When $\alpha>0$, it is used to extract high-frequency signals. $\beta \in R$ is used to control the position of the waveform, $\mu$ is used to scale the size of the extracted waveform. Given a graph signal filter function $f(\lambda)$ of GNN, we multiply $g(\lambda)$ with it to obtain a scaled new graph signal filter function $f^{'}(\lambda)=g(\lambda)\cdot f(\lambda)$. Then we use the monomial bases to approximate the new graph signal filter function $f^{'}(\lambda)$:
\begin{equation}
f^{''}(\lambda)=\sum_{i=0}^{n}\alpha_i\lambda^i
\end{equation}
where $\alpha_i$ is a learnable weight parameter and $n$ represents the number of layers of the polynomial. We randomly sample some points $\mathbf{X}$ in the eigenvalue interval [-1,1] as training samples, and then input them into $f^{'}(\lambda)$ and $f^{''}(\lambda)$, and calculate the Euclidean distance between them as the loss function to train the coefficients $\alpha_i$ of the polynomial. To prevent interference between multi-objective optimization, we pre-train the coefficients of the graph signal filter function:
\begin{equation}
loss_{filter} = ||f^{'}(X)-f^{''}(X)||_{2}
\end{equation}

\begin{figure}[t]
\centering
\begin{minipage}{0.25\textwidth}
  \centering
  \includegraphics[width=1\linewidth]{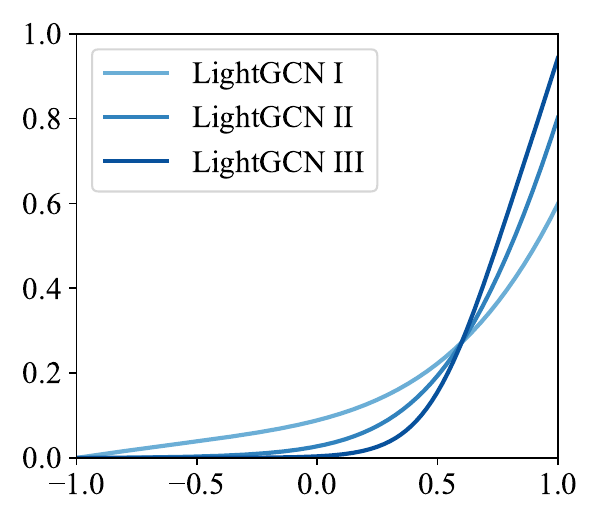}
\end{minipage}%
\centering
\begin{minipage}{0.25\textwidth}
  \centering
  \includegraphics[width=1\linewidth]{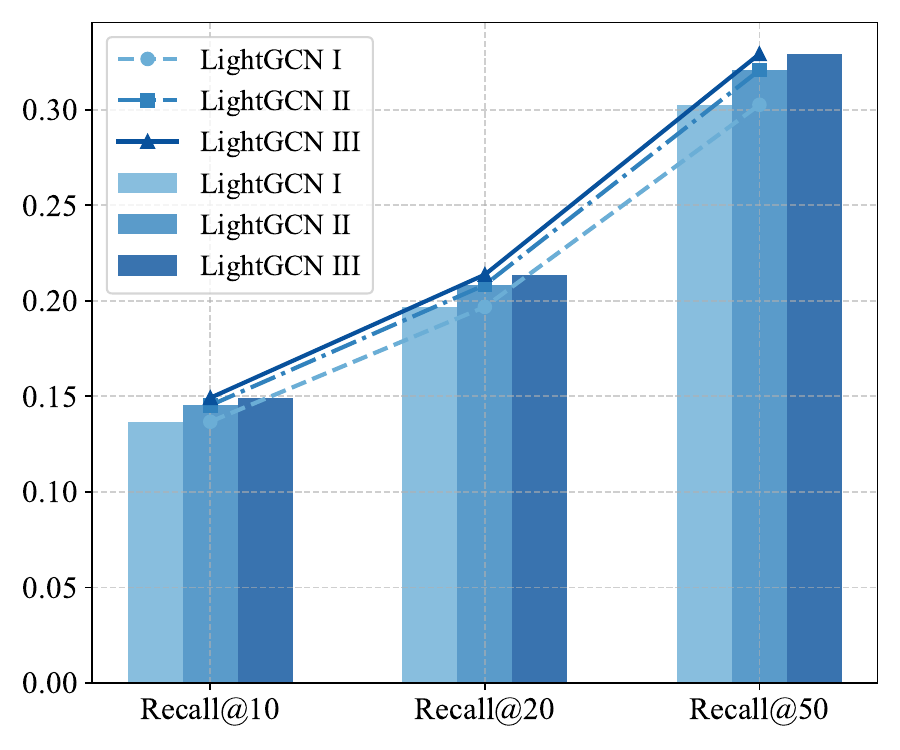}
\end{minipage}
\caption{The performance of LightGCN with different waveforms on Gowalla.}
\label{fig:lgn}
\Description{ }
\end{figure}

Using LightGCN as the base model, we apply the frequency signal scaler to it to adjust the waveform of its filter function. The experimental results are shown in Figure \ref{fig:lgn}.  The performance of LightGCN with different waveforms varies greatly, which is because the waveform of the filter function directly affects the similarity of user-item pairs.

\section{Method}
\begin{figure}[ht]
    \centering
    \includegraphics[width=1\linewidth]{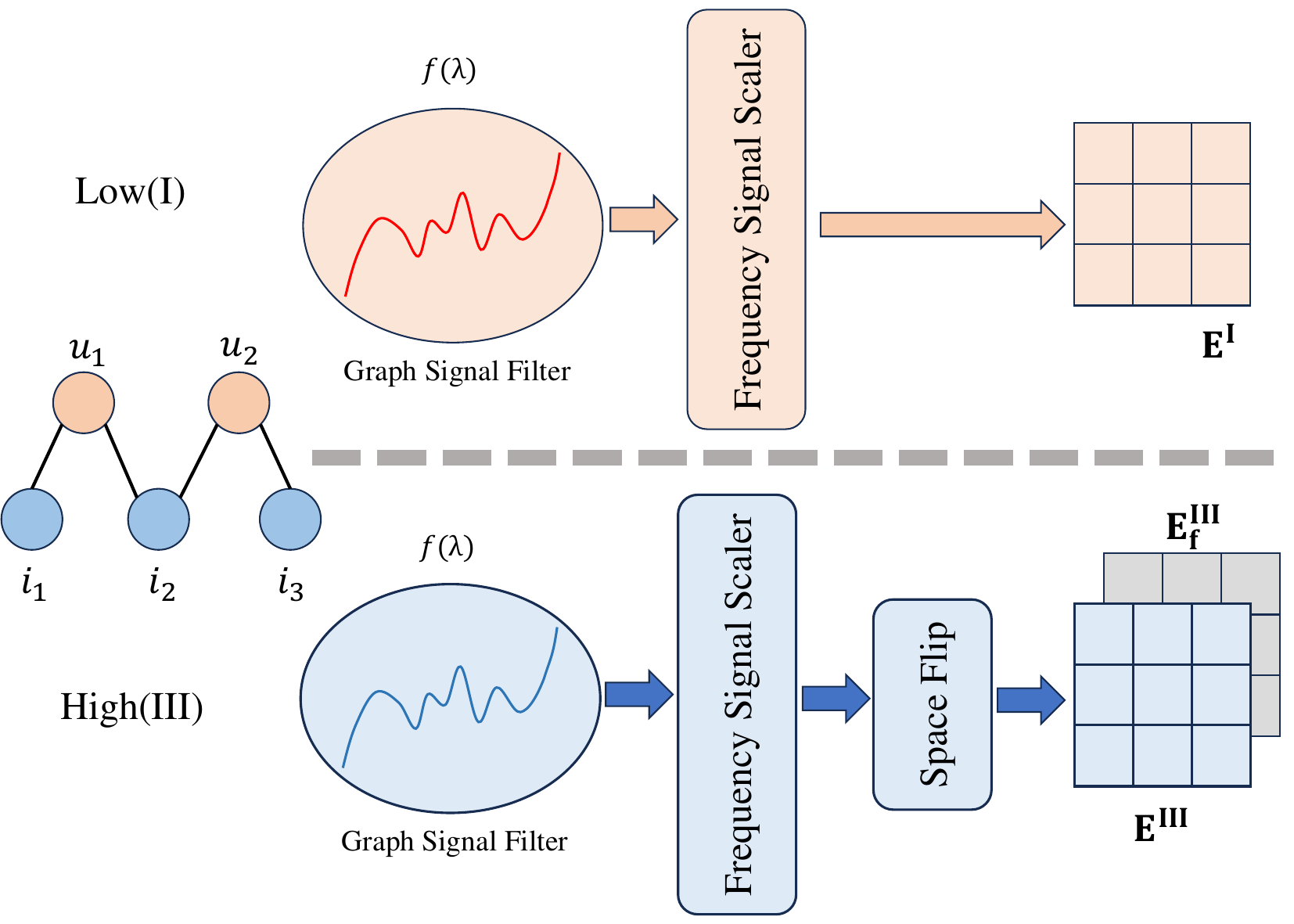}
    \caption{The framework of SimGCF. The low-frequency SimGCF performs frequency signal scaling on the original $f(\lambda)$, and the high-frequency SimGCF also needs to perform space flip on the embedding after scaling.}
    \label{fig:model}
    \Description{ }
\end{figure}

We propose a \textbf{Sim}ple spectral \textbf{G}raph \textbf{C}ollaborative \textbf{F}iltering model \textbf{SimGCF}. As shown in Figure \ref{fig:model}, SimGCF can be divided into two versions: low-pass filtering and high-pass filtering. First, we select a GNN graph signal filter function as the backbone, and then we use the frequency signal scaler to adjust it more finely. In the high-pass filter version, we also need to perform space flip on it. 
\subsection{Backbone Graph Signal Filtering Function}
The common graph signal filter function $f(\lambda)$ of the GNN-based recommendation model is a filter function in the form of a monomial bases, such as LightGCN \cite{he2020lightgcn}. A study attempt to use a filter function based on Jacobi polynomial bases for graph collaborative filtering \cite{guo2023manipulating}, which is in the following form:
\begin{equation}
f(\lambda) = \frac{1}{n+1}\sum_{i=0}^{n}\mathbf{J}_{i}^{a,b}(\lambda)
\end{equation}
where $\mathbf{J}_{i}^{a,b}(\lambda)$ represents the $i$-th order Jacobi polynomial bases of $\lambda$. We choose Jacobi polynomial bases functions as our backbone graph signal filtering functions. In fact, our model is not limited to the Jacobi polynomial filter function, it can be replaced by any other form of filter function.

\subsection{Frequency Signal Scaling}
As described in \ref{lab:FSS}, the waveform of $f(\lambda)$ directly affects the user-item similarity and has an important impact on the recommendation. However, the original polynomial filter function is still not flexible enough to adjust the graph signal, so we propose a frequency signal scaler $g(\lambda)$ to adjust it. The adjusted graph signal filter function $f^{'}(\lambda)$ is as follows:
\begin{equation}
f^{'}(\lambda)=g(\lambda)\cdot f(\lambda)
\end{equation}
We use a monomial bases, $f^{''}(\lambda)=\sum_{i=0}^{n}\alpha_i\lambda^i$, to approximate the adjusted filter function $f^{'}(\lambda)$ to learn the coefficients $\alpha_i$. It can also be replaced by other polynomial bases. Then we express the original graph signal through graph embedding:
\begin{equation}
\begin{aligned}
\mathbf{E}
&=\mathbf{U}f^{''}(\Lambda)\mathbf{U^T}\mathbf{E}^{0}=f^{''}(\mathbf{\hat{A}})\mathbf{E}^{0}=\sum_{i=0}^{n}\alpha_i\mathbf{\hat{A}}^{i}\mathbf{E}^0
\end{aligned}
\end{equation}
Due to the low-frequency GES can directly express the GS of Low(I), there is no need to perform space flip, and $\mathbf{E}^{\mathbf{I}}$ can be obtained directly. However, High(III) GS cannot be directly expressed through GES, and the original high-frequency graph embedding $\mathbf{E}^{\mathbf{III}}$ needs to be spatially flipped $\mathbf{E}^{\mathbf{III}}_{\mathbf{f}}=-\mathbf{E}^{\mathbf{III}}$.

\subsection{Optimization}
We use the inner product of graph embedding to represent the interest score $\hat{y}_{ui}$ of user $u$ and item $i$. On low-frequency SimGCF, $\hat{y}_{ui}=\mathbf{e}^{\mathbf{I}}_{u}*(\mathbf{e}^{\mathbf{ I}}_{i})^{T}$, where $\mathbf{e}^{\mathbf{I}}_{u},\mathbf{e}^{\mathbf{I}}_{i }\in \mathbf{E}^{\mathbf{I}}$. On high-frequency SimGCF, $\hat{y}_{ui}=\mathbf{e}^{\mathbf{f}}_{u}*(\mathbf{e}^{\mathbf{III}} _{i})^{T}$, where $\mathbf{e}^{\mathbf{f}}_{u}\in\mathbf{E}^{\mathbf{III}}_{\mathbf{f}},\mathbf{e}^{\mathbf{III}}_{i} \in\mathbf{E}^{\mathbf{III}}$. Then we use Bayesian Personalized Rank (BPR) \cite{rendle2012bpr} as the recommendation loss:
\begin{equation}
\mathcal{L}_{bpr}=-\sum_{u\in\mathcal{U}}\sum_{i\in\mathcal{N}_u}\sum_{j\notin \mathcal{N}_u}\mathbf{ln}\sigma(\hat{y}_{ui}-\hat{y}_{uj})
\end{equation}
where $\mathcal{N}_u$ is the set of neighbors of user $u$. The final loss function can be expressed as follows:
\begin{equation}
\mathcal{L}=\mathcal{L}_{bpr}+w\lVert\mathbf{E^{0}}\rVert^{2}
\end{equation}
$w$ controls the influence of the $L_2$-regularization term.

\section{Experiments}

\begin{table*}[ht]
    \caption{Main experimental results of SimGCF and baselines. The best result is bolded, and the runner-up is underlined.}
    \label{tab:over-per}
    \centering
    \resizebox{1\textwidth}{!}{
    \begin{tabular}{c|c|ccccccccc|cc}
    \toprule
        Dataset & Metric & BPR & NeuMF & NGCF & DGCF & LightGCN & GTN & RGCF & DirectAU & JGCF$^{*}$ & \textbf{SimGCF(I)} & \textbf{SimGCF(III)} \\ \midrule
        \multirow{4}*{\textbf{Gowalla}} 
        & Recall@10 & 0.1159 & 0.0975 & 0.1119 & 0.1252 & 0.1382 & 0.1403 & 0.1335 & 0.1394 & \underline{0.1515} & \textbf{0.1575} & 0.1557\\ 
        ~ & NDCG@10 & 0.0811 & 0.0664 & 0.0787 & 0.0902 & 0.1003 & 0.1009 & 0.0905 & 0.0991 & \underline{0.1106} & \textbf{0.1136} & 0.1119\\ 
        ~ & Recall@20 & 0.1686 & 0.1470 & 0.1633 & 0.1829 & 0.1983 & 0.2016 & 0.1934 & 0.2014 & \underline{0.2172} & \textbf{0.2222} & 0.2183\\ 
        ~ & NDCG@20 & 0.0965 & 0.0808 & 0.0937 & 0.1066 & 0.1175 & 0.1184 & 0.1081 & 0.1170 & \underline{0.1293} & \textbf{0.1321} & 0.1298\\ 
        \midrule
        \multirow{4}*{\textbf{Amazon-Books}} 
        & Recall@10 & 0.0477 & 0.0342 & 0.0475 & 0.0565 & 0.0620 & 0.0588 & 0.0712 & 0.0683 & \underline{0.0746} & \textbf{0.0803} & 0.0792\\ 
        ~ & NDCG@10 & 0.0379 & 0.0266 & 0.0330 & 0.0448 & 0.0506 & 0.0485 & 0.0568 & 0.0569 & \underline{0.0624} & \textbf{0.0673} & 0.0663\\ 
        ~ & Recall@20 & 0.0764 & 0.0575 & 0.076 & 0.0867 & 0.0953 & 0.0930 & 0.1090 & 0.1053 & \underline{0.1144} & \textbf{0.1202} & 0.1196\\ 
        ~ & NDCG@20 & 0.0474 & 0.0345 & 0.0472 & 0.0551 & 0.0615 & 0.0597 & 0.0697 & 0.0689 & \underline{0.0754} & \textbf{0.0801} & 0.0794\\ 
        \midrule
        \multirow{4}*{\textbf{Yelp2018}} 
        & Recall@10 & 0.0452 & 0.0313 & 0.0459 & 0.0527 & 0.0560 & 0.0603 & 0.0633 & 0.0557 & \underline{0.0669} & \textbf{0.0679} & 0.0672\\ 
        ~ & NDCG@10 & 0.0355 & 0.0235 & 0.0364 & 0.0419 & 0.0450 & 0.0483 & 0.0503 & 0.0435 & \underline{0.0541} & \textbf{0.0550} & 0.0549\\
        ~ & Recall@20 & 0.0764 & 0.0548 & 0.0778 & 0.0856 & 0.0913 & 0.0984 & 0.1026 & 0.0907 & \underline{0.1066} & 0.1075 & \textbf{0.1082}\\
        ~ & NDCG@20 & 0.046 & 0.0316 & 0.0472 & 0.0528 & 0.0569 & 0.0611 & 0.0637 & 0.0553 & \underline{0.0673} & 0.0682 & \textbf{0.0686}\\ 
        \midrule
        \multirow{4}*{\textbf{Alibaba-iFashion}} 
        & Recall@10 & 0.0303 & 0.0182 & 0.0382 & 0.0447 & 0.0477 & 0.0406 & 0.0450 & 0.0319 & \underline{0.0559} & \textbf{0.0571} & 0.0570\\ 
        ~ & NDCG@10 & 0.0161 & 0.0092 & 0.0198 & 0.0241 & 0.0255 & 0.0217 & 0.0249 & 0.0166 & \underline{0.0303} & 0.0310 & \textbf{0.0311}\\ 
        ~ & Recall@20 & 0.0467 & 0.0302 & 0.0615 & 0.0677 & 0.0720 & 0.0625 & 0.0674 & 0.0484 & \underline{0.0823} & \textbf{0.0848} & 0.0838\\ 
        ~ & NDCG@20 & 0.0203 & 0.0123 & 0.0257 & 0.0299 & 0.0316 & 0.0272 & 0.0305 & 0.0207 & \underline{0.0370} & \textbf{0.0380} & 0.0379\\ 
        \bottomrule
    \end{tabular}
    }
    \begin{minipage}{0.95\textwidth}
        \footnotesize
        \textbf{Note:} $^{*}$ means that only low-frequency and high-frequency graph signals are considered, and mid-frequency signals are removed.
    \end{minipage}
\end{table*}

We conduct experiments to answer the following questions: \textbf{Q1}: How does SimGCF compare to other baselines? \textbf{Q2}: What are the effects of space flip and frequency signals scaler in SimGCF? \textbf{Q3}: What is the impact of different hyper-parameters in SimGCF? \textbf{Q4}: What is the time complexity of SimGCF?
\begin{table}[H]
    \caption{Statistics of the Datasets.}
    \label{tab:dataset}
    \centering
    \resizebox{1\linewidth}{!}{
    \begin{tabular}{lllll}
    \toprule
        \textbf{Datasets} & \textbf{Users} & \textbf{Items} & \textbf{Interactions} & \textbf{Sparsity} \\ 
        \midrule
        Gowalla & 29,858 & 40,981 & 1,027,464 & 99.92\%\\ 
        Amazon-Books & 52,642 & 91,598 & 2,984,108 & 99.94\% \\ 
        Yelp2018 & 31,668 & 38,048 & 1,561,406 & 99.87\% \\ 
        Alibaba-iFashion & 300,000 & 81,614 & 1,607,813 & 99.99\% \\ 
    \bottomrule
    \end{tabular}
    }
\end{table}

\subsection{Experimental Setup}
\subsubsection{Datasets.}
\label{adjust_dataset}
We conduct experiments on four widely used public datasets. Gowalla \cite{cho2011friendship}, Amazon-Books \cite{mcauley2015image}, Yelp \cite{wang2019neural} and Alibaba-iFashion \cite{chen2019pog}. The details of the datasets are shown in Table \ref{tab:dataset}. We randomly sample 80\% of the user interaction data for training, 10\% of the data for validation, and the remaining 10\% for testing. 

\subsubsection{Baselines.}
In our evaluation, SimGCF is benchmarked against a series of established baselines in the field. We follow the experimental setup outlined by Guo et al. \cite{guo2023manipulating}, including hyperparameters and training protocols, to ensure consistency and comparability with established benchmarks. The baseline models include:
\begin{itemize}
    \item \textbf{BPR} \cite{rendle2012bpr} is a matrix factorization framework based on Bayesian empirical loss. 
    \item \textbf{NeuMF} \cite{he2017neural} replaces the dot product in the MF model with a multi-layer perceptron to learn the match function of users and items. 
    \item \textbf{NGCF} \cite{wang2019neural} adopts user-item bipartite graphs to incorporate high-order information and utilizes GNN to enhance CF. 
    \item \textbf{DGCF} \cite{wang2020disentangled} produces disentangled representations for the user and item to improve the performance. 
    \item \textbf{LightGCN} \cite{he2020lightgcn} simplifies the design of GCN to make it more concise and appropriate for recommendation. 
    \item \textbf{GTN} \cite{fan2022graph} proposes to use graph trend filtering to denoise the user-item graph. 
    \item \textbf{RGCF} \cite{tian2022learning} learns to denoise the user-item graph by removing noisy edges and then adding edges to ensure diversity. 
    \item \textbf{DirectAU} \cite{wang2022towards} is a collaborative filtering method with representation alignment and uniformity. 
    \item \textbf{JGCF} \cite{guo2023manipulating} is a spectral GNN based on Jacobi polynomial sets that can effectively utilize graph signals.
\end{itemize}

\subsubsection{Evaluation Metrics.}
We use two widely used metrics, Recall@$k$ and NDCG@$k$, to evaluate the performance of top-$k$ recommendation. In our experiments, we set $k$ to 10, 20 for reporting.

\subsubsection{Implementation Details.}
Our experiments are based on the RecBole framework \cite{xu2023towards}. We optimize all baselines with Adam and carefully select hyperparameters according to their suggestions. The batch size is set to 4096 for Gowalla, Yelp, and Alibaba-iFashion, and the batch size is set to 8192 for Amazon-Books. All parameters are initialized by Xavier distribution. The embedding size of all methods is set to 64. Early stopping of 5 epochs is used to prevent overfitting, with Recall@20 as the metric. We use the parameter configuration of JGCF provided by the original paper. The experiments are based on a 32GB Tesla V100-PCIE GPU.

\subsection{Overall Performance (\textbf{Q1})}
We use Low(I) and High(III) with the same absolute value as the filter function of SimGCF to obtain SimGCF(I) and SimGCF(III). The experimental results are shown in Table \ref{tab:over-per}. From these results, we can draw the following conclusions:

Firstly, the GNN-based methods outperform traditional matrix factorization-based methods. The main reason is that the GNN-based model can utilize high-order information on the graph, compared to MF which can only use first-order interaction information for model training. Some recent work on denoising in GNN (GTN and RGCF) has achieved good results on most datasets. They further improve the performance of the model compared to LightGCN by eliminating the noise information in the interaction behavior. DirectAU achieves impressive performance by proposing a new loss for optimizing the uniformity and alignment of representations in recommendations. The above graph models are all based on GCN in the spatial domain and have poor processing capabilities for graph signals. JGCF is a spectral GNN model based on the Jacobi polynomial bases. It controls the waveform of the graph signal filter by adjusting the values of hyperparameters a and b. It can more effectively learn signals of various frequencies on the graph and has demonstrated outstanding performance.

Moreover, SimGCF(I) consistently outperforms all baselines on all public datasets, which demonstrates the effectiveness of our proposed graph signal scaler. Due to the target of this paper is low-frequency and high-frequency graph signals, we choose to remove the mid-frequency signals in JGCF to eliminate the interference of mid-frequency signals. By comparing JGCF and SimGCF (I), we can find that it is beneficial to adjust the frequency signal of the original JGCF waveform. This is because the waveform of the graph signal filter function directly affects the similarity of user-item pairs in the recommendation. In fact, we find that the high-frequency signal in JGCF hinders the model's performance, which can be mitigated by filtering it out using a frequency signal scaler. This will be discussed in detail in the next section.

Finally, the performances of SimGCF(I) and SimGCF(III) are very close on all datasets. This demonstrates that the low-frequency graph signal Low (I) and the high-frequency graph signal High (III) with the same absolute value are equivalent in recommendation. Both influence the performance of the final recommendation by modulating the similarity between user-item pairs. Due to the limitations of current GES in capturing high-frequency GS beneficial for recommendation, we propose the space flip method to restores these signals through GES. 

\subsection{Ablation Studies (\textbf{Q2})}
To study the effects of space flip (SF) and frequency signal scaler (FSS) in SimGCF, we construct the following variants: JGCF, JGCF(H) and JGCF(L) represent the original JGCF, low-frequency JGCF and high-frequency JGCF respectively. JGCF(H)+SF means using SF for high-frequency JGCF. SimGCF(I) means performing FSS on the low-frequency JGCF, and SimGCF(III) means performing FSS and SF on the high-frequency JGCF.
\begin{table}[h]
    \centering
    \caption{Ablation study of SimGCF. We abbreviate Recall and NDCG as R and N respectively.}
    \label{tab:ablation}
    \resizebox{1\linewidth}{!}{
    \begin{tabular}{c|cc|cc|cc|cc}
        \toprule
        \multirow{2}*{Variants}
        & \multicolumn{2}{c|}{\textbf{Gowalla}} & \multicolumn{2}{c|}{\textbf{Amazon-Books}} & \multicolumn{2}{c|}{\textbf{Yelp2018}} & \multicolumn{2}{c}{\textbf{Alibaba-iFasion}}\\
        & R@20 & N@20 & R@20 & N@20 & R@20 & N@20 & R@20 & N@20 \\
        \midrule
        JGCF & 0.2172 & 0.1293 & 0.1144 & 0.0754 & 0.1066 & 0.0673 & 0.0823 & 0.0370 \\
        JGCF(H) & 0.0383 & 0.0225 & 0.0098 & 0.0058 & 0.0494 & 0.0301 & 0.0222 & 0.0087 \\
        JGCF(H)+SF & 0.1300 & 0.0754 & 0.0404 & 0.025 & 0.0522 & 0.0313 & 0.0639 & 0.0364 \\
        JGCF(L) & 0.2212 & 0.1301 & 0.1167 & 0.0776 & 0.1068 & 0.0676 & 0.0834 & 0.0376 \\
        \midrule
        \textbf{SimGCF(I)} & 0.2222 & 0.1321 & 0.1202 & 0.0801 & 0.1075 & 0.0682 & 0.0848 & 0.0380 \\
        \textbf{SimGCF(III)} & 0.2183 & 0.1298 & 0.1196 & 0.0794 & 0.1082 & 0.0686 & 0.0838 & 0.0379 \\
        \bottomrule
    \end{tabular}
    }
\end{table}

As shown in Table \ref{tab:ablation}, by comparing JGCF, JGCF(H) and JGCF(L), we can find that the high-frequency signal in the original JGCF suppresses the expressiveness of the model. If we only retain the low-frequency signal, the performance of the model will improve. This is because GES cannot represent the high-frequency graph signals that are beneficial for recommendation. Consequently, the direct fusion of high-frequency and low-frequency signals hinders the model's performance.

Additionally, the comparison between JGCF(H)+SF and JGCF(H) indicates that SF effectively recovers the high-frequency graph signals that are beneficial for recommendation. Meanwhile, the performance comparison between JGCF(L) and SimGCF(I) shows that FSS remains advantageous for adjusting the filter function. Since the original low-frequency waveform of JGCF is already close to optimal, the performance improvement brought by FSS on JGCF is less significant than that on LightGCN. Lastly, the performance of SimGCF(I) is very similar to SimGCF(III), further confirming that high-frequency and low-frequency signals with the same absolute waveform are equivalent in recommendation systems.

\subsection{Sensitivity Analysis (\textbf{Q3})}
\begin{figure}[h]
\centering
\includegraphics[width=1\linewidth]{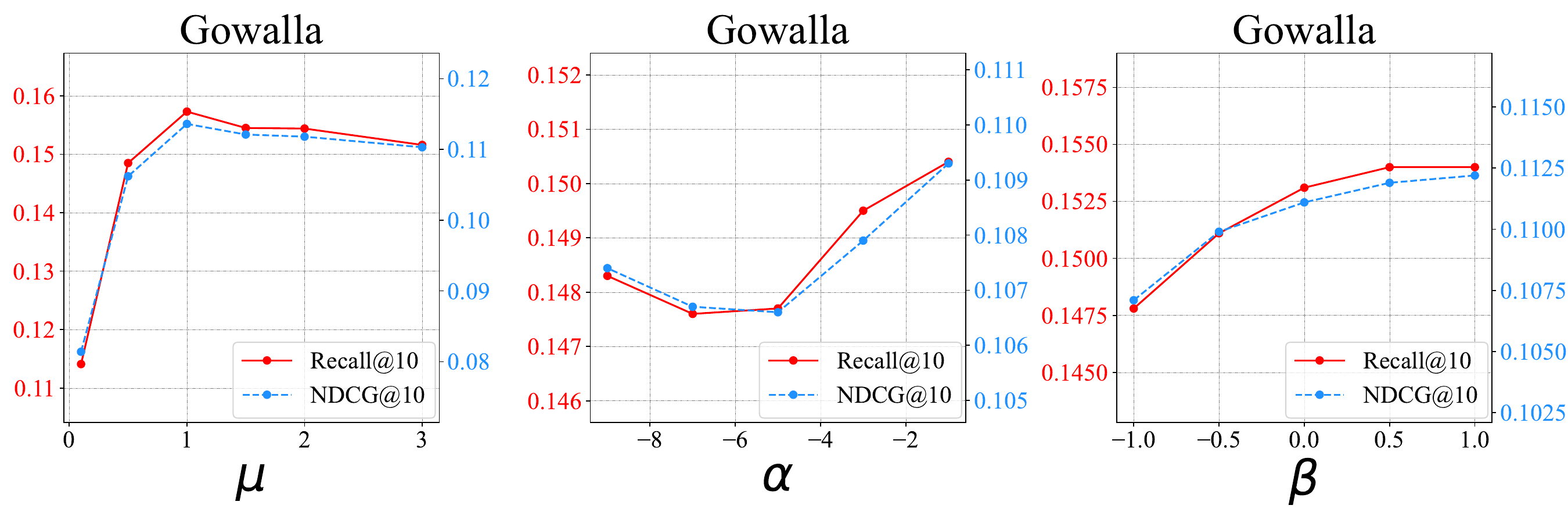}
\includegraphics[width=1\linewidth]{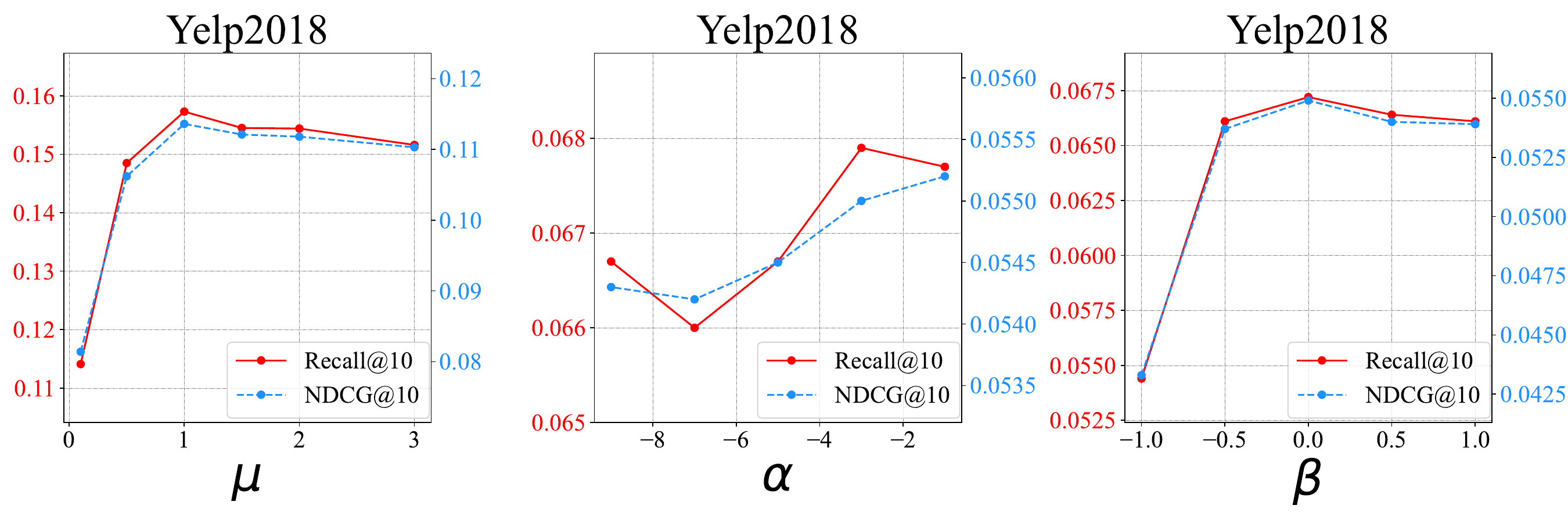}
\includegraphics[width=1\linewidth]{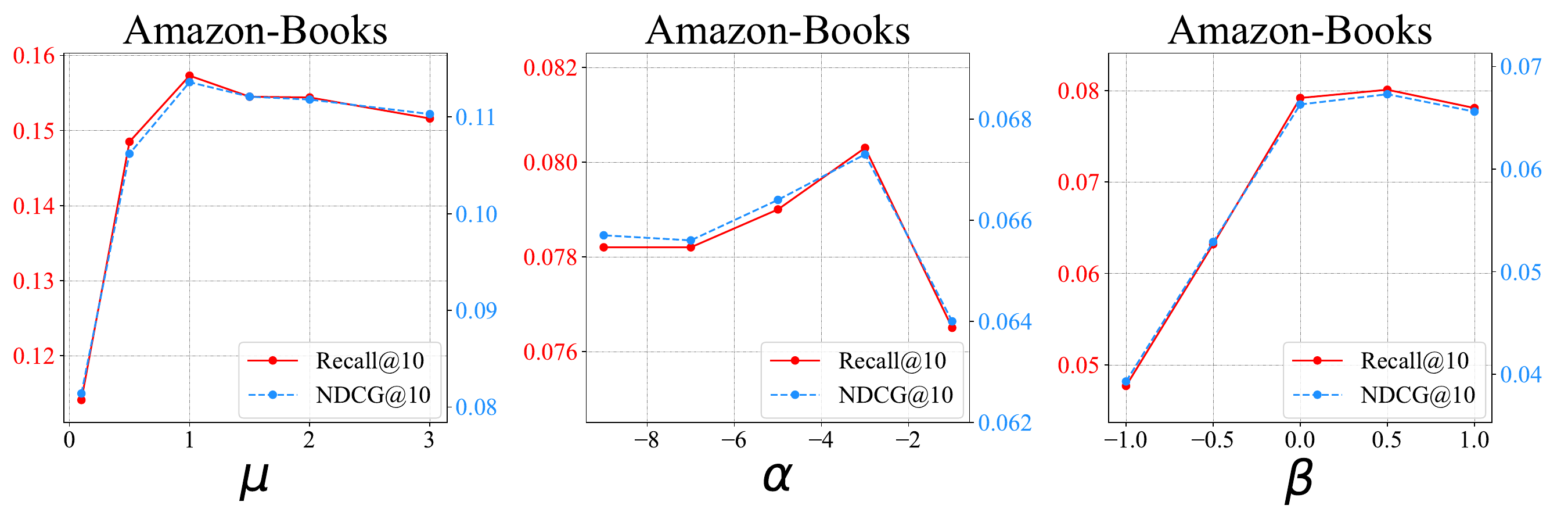}
\includegraphics[width=1\linewidth]{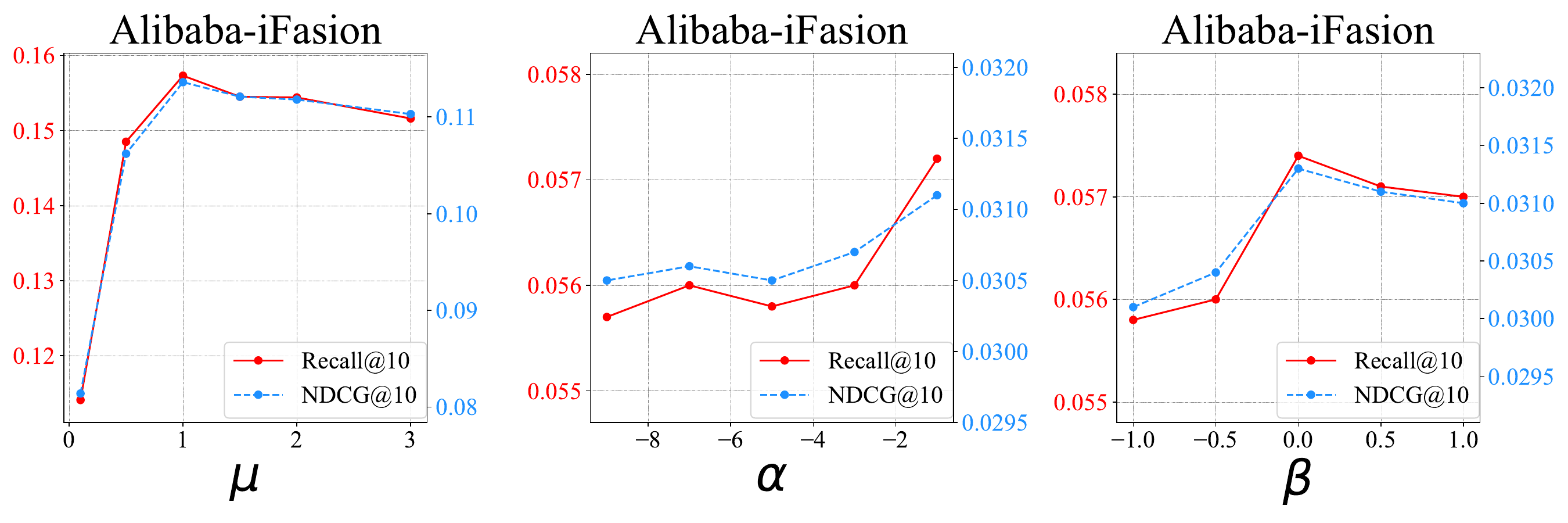}
\caption{Analysis of $\mu$, $\alpha$ and $\beta$.}
\label{fig:params}
\Description{ }
\end{figure}

We take SimGCF(I) as an example to perform parameter sensitivity analysis. The results of parameters $\mu$, $\alpha$, and $\beta$ are shown in Figure \ref{fig:params}. On the four datasets, we observe that the value of $\mu$ around 1 achieved the best performance. $\mu$ controls the scaling of the absolute value of $f(\lambda)$, and we observe that when $\mu<1$, the performance of the model increases as the value of $\mu$ increases. When $\mu>1$, the performance of the model decreases as the value of $\mu$ increases. This indicates that the waveform of the original graph signal filter function does not need to be scaled to a great extent.

The parameter $\alpha$ controls the steepness of the waveform of $f(\lambda)$, which has a greater impact on the performance of the model. We can observe that on Yelp2018 and Amazon-Books, the value of $\alpha$ reaches the optimal value around -3. When $\alpha<-3$, the performance of the model increases as the value of $\alpha$ increases. When $\alpha>-3$, the performance decreases as the value of $\alpha$ increases. On Alibaba-iFasion, $\alpha=-1$ reaches the optimal value. We observe that the steepness of the graph signal filter function has a significant impact on model performance. By adjusting the steepness of the original $f(\lambda)$, the similarity between user-item pairs is directly influenced.

The parameter $\beta$ controls the adjustment range of the frequency signal scaler, which in turn significantly affects the model's overall performance. On the Gowalla, Yelp2018, Amazon-Books, and Alibaba-iFashion datasets, the best results are obtained when $\beta$ is set to 0.5, 0, 0.5, and 0, respectively.

\subsection{Complexity Analysis (\textbf{Q4})}
\begin{figure}[ht]
    \centering
    \includegraphics[width=1\linewidth]{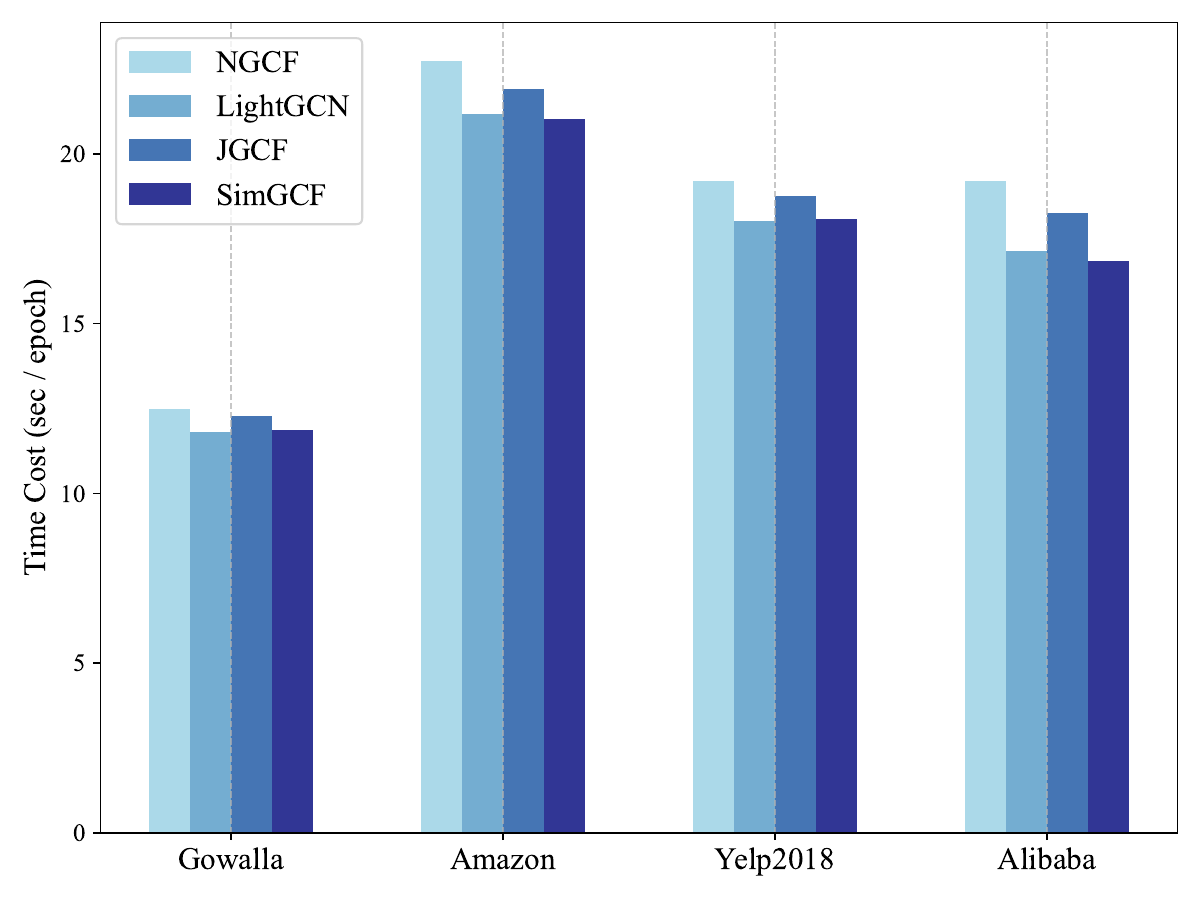}
    \caption{The time cost on each training epoch.}
    \label{fig:time}
    \Description{ }
\end{figure}

We explore the time complexity of SimGCF. SimGCF consists of two stages. The first stage involves the pre-training of the graph signal filtering function. The first stage uses a polynomial basis to approximate the filtering function and learn its coefficients, requiring no complex matrix operations and incurring negligible computational cost. In the second stage, we perform spectral graph convolution with the pre-trained coefficients. Since this stage involves matrix operations, we compare the per-epoch time complexity of SimGCF with other models. As shown in Figure \ref{fig:time}, the overall efficiency of SimGCF is close to that of LightGCN, and is better than JGCF and NGCF. In fact, LightGCN can be considered a special case of SimGCF, as SimGCF employs the monomial bases.

\section{Conclusion}
In this work, we investigate the influence of graph signals on recommender systems. We reveal and theoretically demonstrate that low-frequency and high-frequency graph signal filters with the same absolute value are equivalent in recommendation. They both affect recommendation by smoothing the similarities between user-item pairs. Furthermore, we identify that existing graph embedding methods can only express half of the characteristics of the graph signal. To address this, we propose SimGCF, a spectral GNN model with enhanced representation capabilities, offering greater flexibility in utilizing graph structure information. Extensive theoretical analysis and experiments validate the effectiveness of SimGCF.

\begin{acks}
This research was partially supported by the NSFC (62376180, 62506255, 62176175, 62176014),  the Priority Academic Program Development of Jiangsu Higher Education Institutions and the Fundamental Research Funds for the Central Universities.
\end{acks}

\clearpage
\bibliographystyle{ACM-Reference-Format}
\balance
\bibliography{SIGKDD2026}

\appendix
\section{GRAPH SIGNALS AND GRAPH EMBEDDING SIGNALS}
\label{lab:gs_ges}
\begin{figure}[h]
    \centering
    \includegraphics[width=1\linewidth]{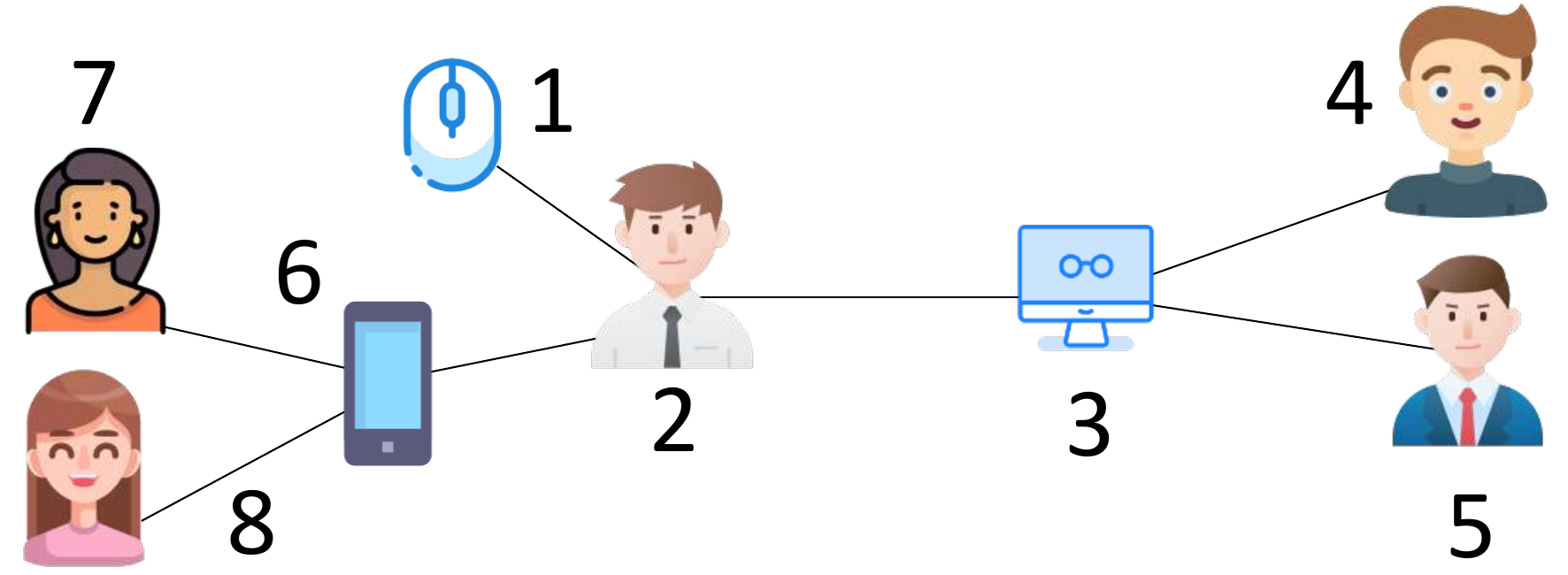}
    \caption{The case graph.}
    \label{fig:graph}
    \Description{ }
\end{figure}
\begin{figure}[ht]
    \centering
    \includegraphics[width=0.55\linewidth]{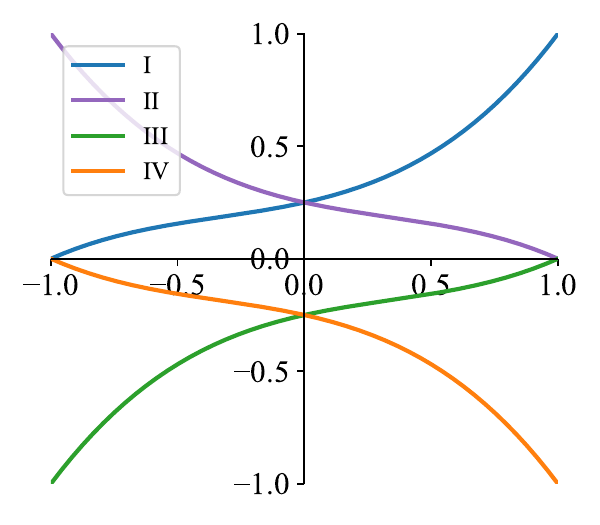}
    \caption{The waveforms of the graph signal filter function in four quadrants.}
    \label{fig:filter}
    \Description{ }
\end{figure}

This section aims to explore the characteristics and differences between graph signals (GS) and graph embedding signals (GES). For visualization, we construct a small-scale case diagram $\mathbf{A}$. 
Without loss of generality, our results can be extended to large-scale graphs. 
As shown in Figure \ref{fig:graph}, the graph contains 5 users and 3 products. We use four graph signal filter functions $f_{\mathbf{I}}(\lambda)$, $f_{\mathbf{II}}(\lambda)$, $f_{\mathbf{III}}(\lambda)$ and $f_{\mathbf{IV}}(\lambda)$ with the same absolute value in different quadrants to perform spectral operations on it, as shown in Figure \ref{fig:filter}. Among them, $f_{\mathbf{I}}(\lambda)$ and $f_{\mathbf{IV}}(\lambda)$ are low-pass filter functions, and $f_{\mathbf{II}(\lambda)}$ and $f_{\mathbf{III}}(\lambda)$ are high-pass filter functions.

\subsection{Graph Signals}
\begin{figure}[h]
\centering
\begin{minipage}{0.25\textwidth}
  \centering
  \includegraphics[width=1\linewidth]{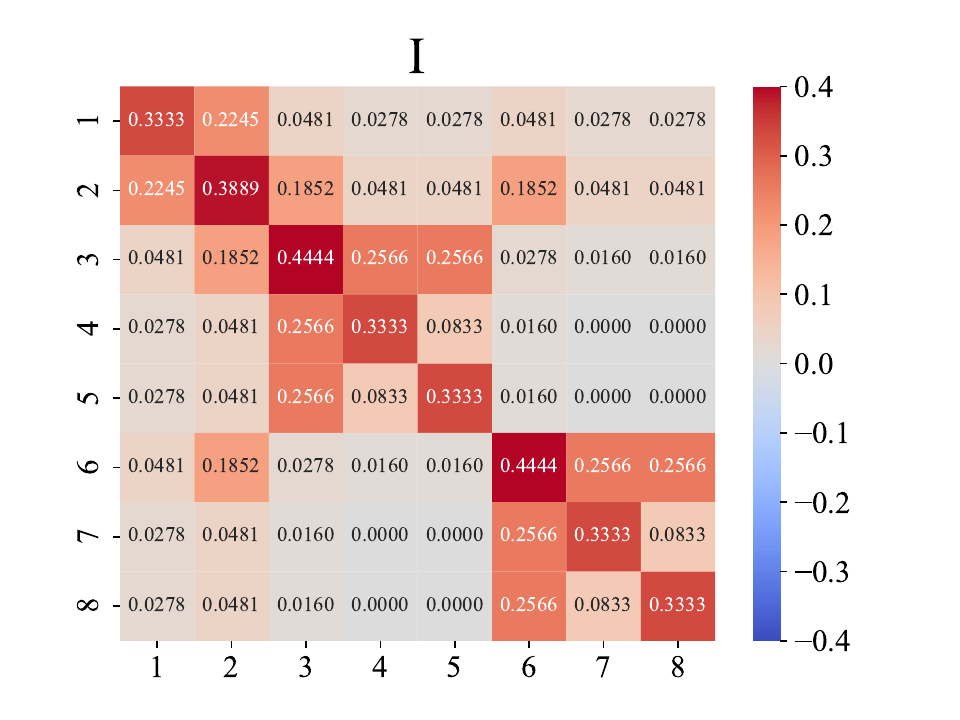}
\end{minipage}%
\begin{minipage}{0.25\textwidth}
  \centering
  \includegraphics[width=1\linewidth]{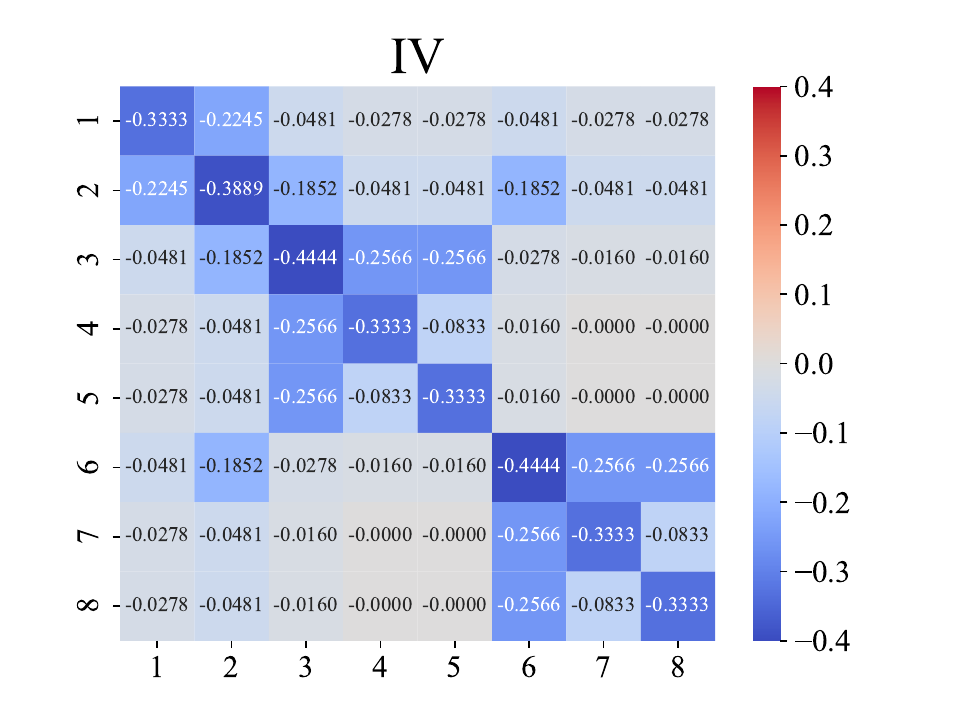}
\end{minipage}
\begin{minipage}{0.25\textwidth}
  \centering
  \includegraphics[width=1\linewidth]{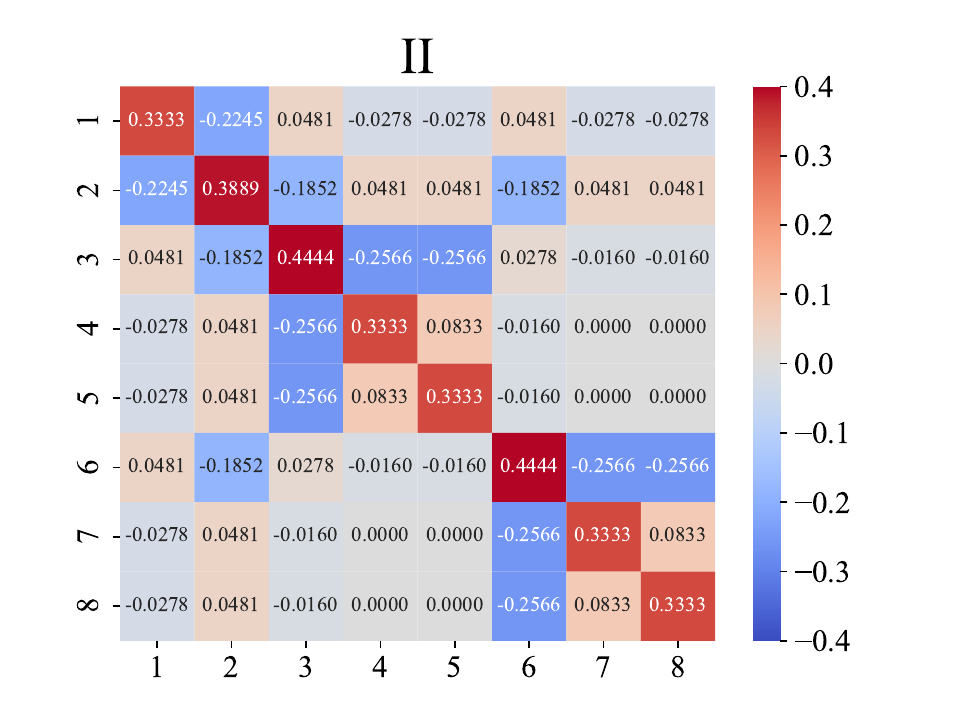}
\end{minipage}%
\begin{minipage}{0.25\textwidth}
  \centering
  \includegraphics[width=1\linewidth]{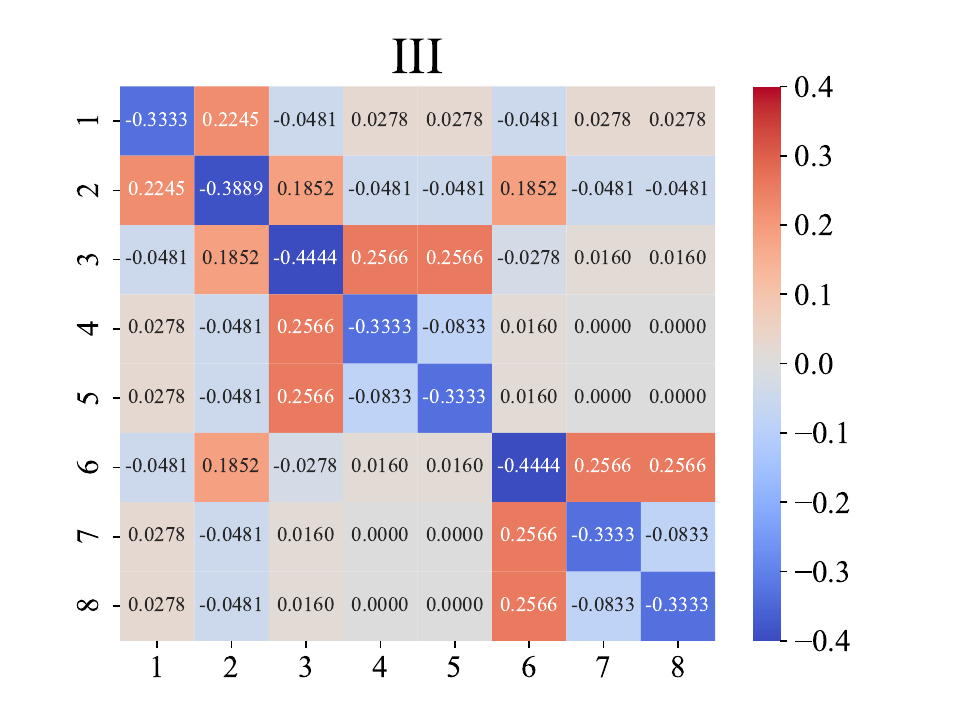}
\end{minipage}
\caption{The characteristics of the graph signals in the four different quadrants.}
\label{fig:adj}
\Description{ }
\end{figure}

The spectral operation of the graph signal is as follows:
\begin{equation}
\mathbf{S}_{1}=f(\mathbf{\hat{A}})=\sum_{i=0}^{n}\alpha_i\mathbf{\hat{A}}
\end{equation}
where $\mathbf{\hat{A}}=\mathbf{D}^{-\frac{1}{2}}\mathbf{A}\mathbf{D}^{-\frac{1}{2}}$ represents the normalized adjacency matrix. $n$ represents the order of the polynomial, which is set to 3 in our experiment. The characteristics of the GS obtained after filtering by the graph signal filter functions of four different quadrants are shown in Figure \ref{fig:adj}.

From the experimental results, we can find that the low-frequency GS $f_{\mathbf{I}}(\lambda)$ and $f_{\mathbf{IV}}(\lambda)$ show the characteristics of stable changes, and the absolute value of the similarity between nodes decays with the increase of the path length. Taking the path 1->2->3->4 as an example, the similarities between node 1 and nodes 2, 3, and 4 are 0.2245, 0.0481, and 0.0278, respectively. When $f(\lambda)>0$, the low-frequency GS $f_{\mathbf{I}}(\lambda)$ makes the connected nodes similar, while when $f(\lambda)<0$, the low-frequency GS $f_{\mathbf{IV}}(\lambda)$ makes the connected nodes dissimilar. The high-frequency graph signals $f_{\mathbf{II}}(\lambda)$ and $f_{\mathbf{III}}(\lambda)$ show the characteristics of frequent changes. When $f(\lambda)>0$, the high-frequency GS $f_{\mathbf{II}}(\lambda)$ will make odd-order nodes dissimilar and even-order nodes similar. As shown in the case graph, the similarities of odd-order nodes 1 and 2, 1 and 4 are -0.2245 and -0.0278 respectively, and the similarities of even-order nodes 2 and 4, 2 and 7 are 0.0481 and 0.0481 respectively. When $f(\lambda)<0$, the high-frequency GS $f_{\mathbf{III}}(\lambda)$ will make odd-order nodes similar and even-order nodes dissimilar.

\subsection{Graph Embedding Signals}
\begin{figure}[ht]
\centering
\begin{minipage}{0.25\textwidth}
  \centering
  \includegraphics[width=1\linewidth]{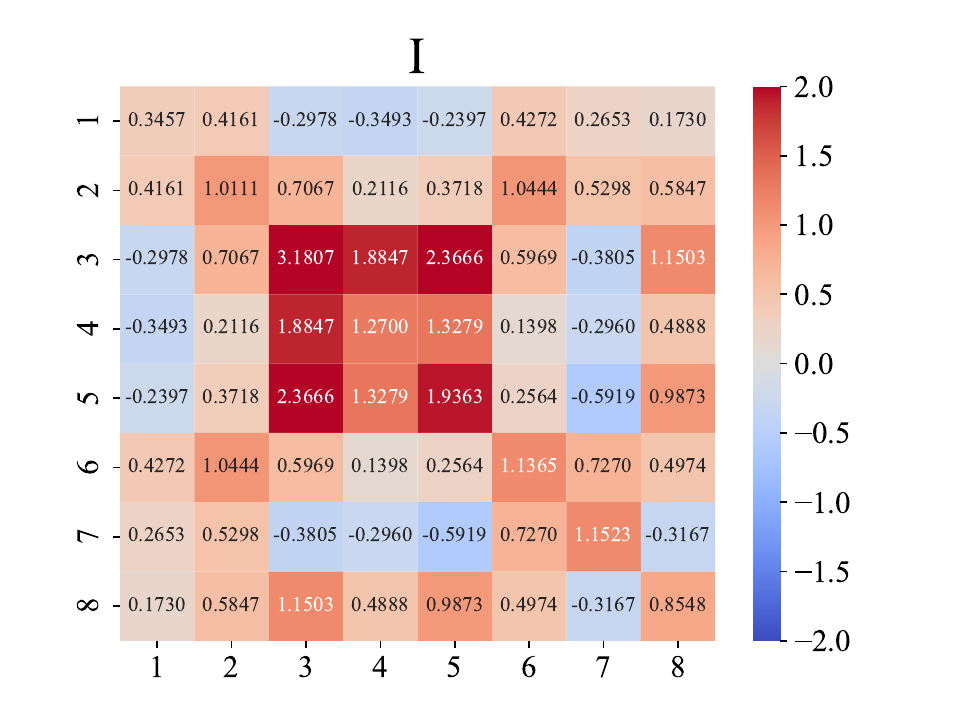}
\end{minipage}%
\begin{minipage}{0.25\textwidth}
  \centering
  \includegraphics[width=1\linewidth]{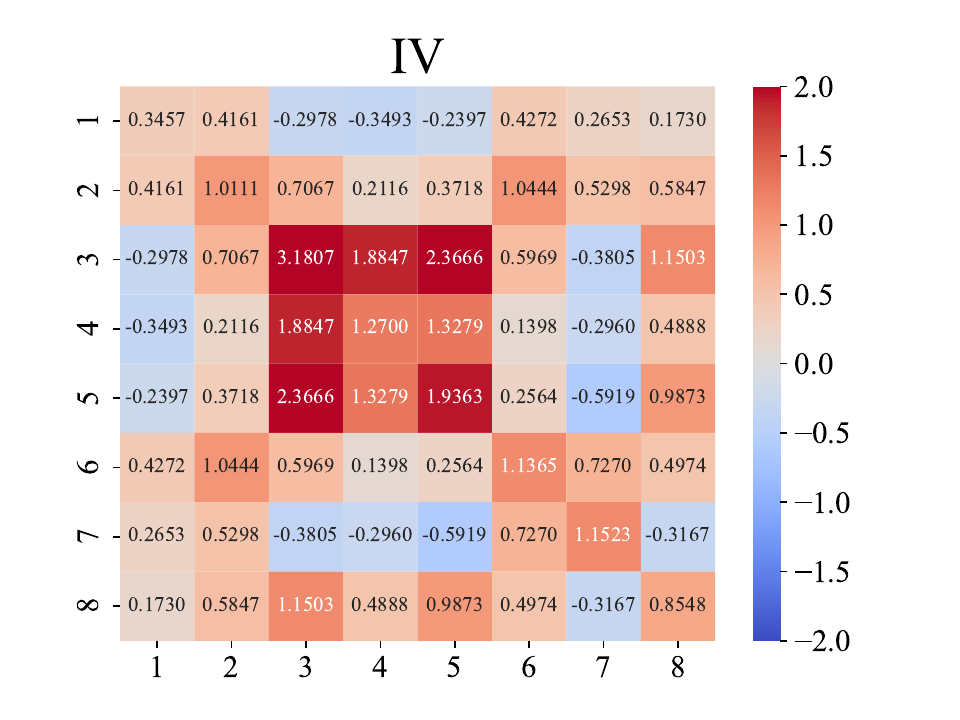}
\end{minipage}
\begin{minipage}{0.25\textwidth}
  \centering
  \includegraphics[width=1\linewidth]{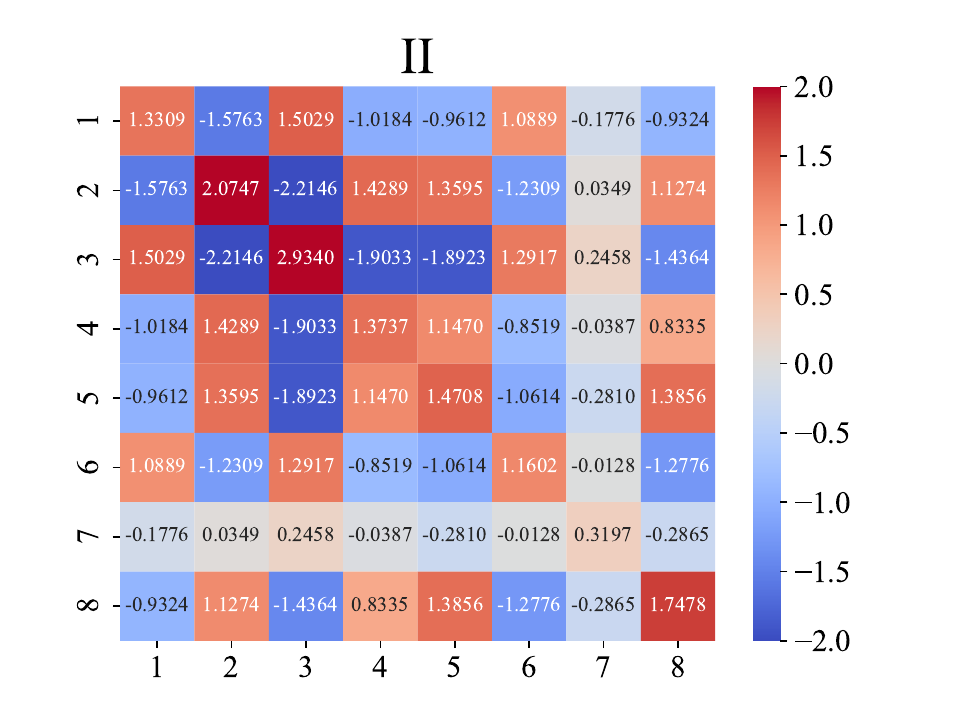}
\end{minipage}%
\begin{minipage}{0.25\textwidth}
  \centering
  \includegraphics[width=1\linewidth]{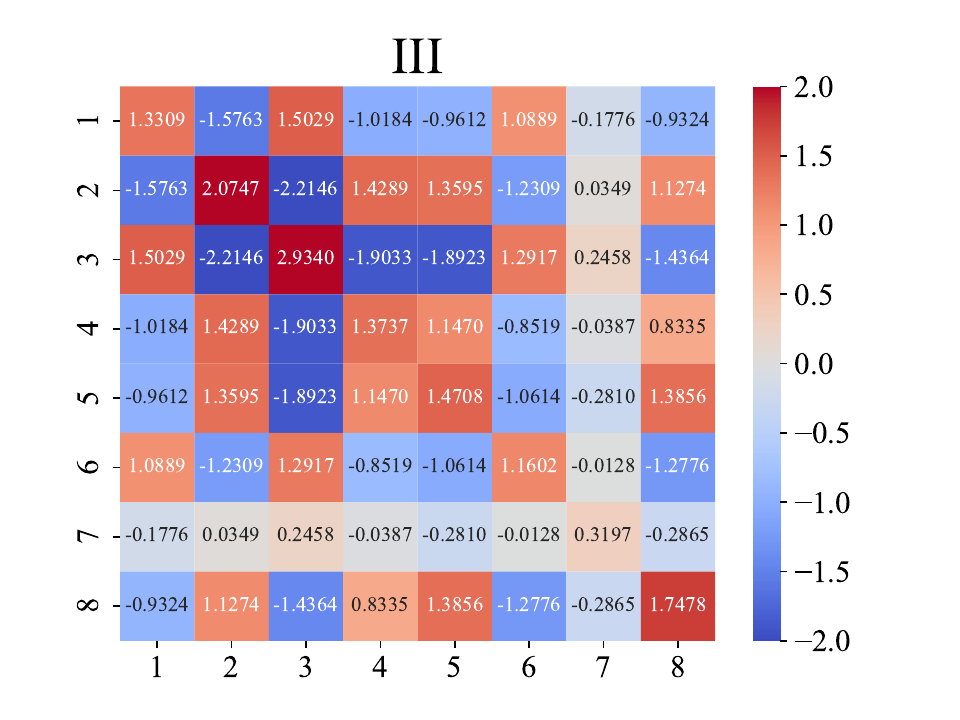}
\end{minipage}
\caption{The characteristics of the graph embedding signals in the four different quadrants.}
\label{fig:emb}
\Description{ }
\end{figure}

The graph embedding signal is in the form of:
\begin{equation}
\mathbf{S}_{2}=\mathbf{U}f(\mathbf{\Lambda})\mathbf{U}^{\mathbf{T}}\mathbf{E}^{0}=f(\mathbf{\hat{A}})\mathbf{E}^{0}=\sum_{i=0}^{n}\alpha_i\mathbf{\hat{A}}^i\mathbf{E}^0
\end{equation}
where $\mathbf{E}^0$ represents the initialized node embedding, and we use random initialization. $n$ is still set to 3. The characteristics of the GES obtained after filtering by the graph signal filter functions in four different quadrants are shown in Figure \ref{fig:emb}.

From the experimental results, we can find that the low-frequency GES $f_{\mathbf{I}}(\lambda)$ and $f_{\mathbf{IV}}(\lambda)$ show the characteristics of stable changes, and the absolute value of the similarity between nodes decays with the increase of the path length. The difference from the GS is that the features of $f_{\mathbf{I}}(\lambda)$ and $f_{\mathbf{IV}}(\lambda)$ of the GES are the same. This is because the GES cannot express the negative sign when calculating the similarity, resulting in the inability to express the feature of $f(\lambda)<0$. Similarly, the high-frequency GES $f_{\mathbf{II}}(\lambda)$ and $f_{\mathbf{III}}(\lambda)$ cannot express the feature of $f(\lambda)<0$.

\section{How DO GRAPH SIGNALS AFFECT RECOMMENDATION?}
\label{lab:gs_in_rec}
Let's take Low(I) and High(III) as examples. Given the filter functions $f_{\mathbf{I}}(\lambda)$ and $f_{\mathbf{III}}(\lambda)$ of Low(I) and High(III), as follows:
\begin{equation}
f_{\mathbf{I}}(\lambda)=\sum_{i=0}^{n}\alpha_i\lambda^i
\end{equation}
\begin{equation}
f_{\mathbf{III}}(\lambda)=\sum_{i=0}^{n}(-1)^{i+1}\alpha_i\lambda^i
\end{equation}
where $n$ represents the number of layers of the polynomial bases. We embed them, $\mathbf{E}_{\mathbf{I}}=f_{\mathbf{I}}(\lambda)\mathbf{E}^0$, $\mathbf{E}_{\mathbf{III}}=f_{\mathbf{III}}(\lambda)\mathbf{E}^0$. Then we calculate the graph embedding signals $\mathbf{S}_{\mathbf{I}}$ and $\mathbf{S}_{\mathbf{III}}$ of Low(I) and High(III) as follows. Note that we need to perform space flip when calculating $\mathbf{S}_{\mathbf{III}}$:
\begin{equation}
\begin{aligned}
\mathbf{S}_{\mathbf{I}}
&=\mathbf{E}_{\mathbf{I}}\mathbf{E}_{\mathbf{I}}^{T}\\
&=\left(
\begin{aligned}
(1)^{0}\alpha_0\alpha_0\mathbf{E}^0(\mathbf{E}^0)^{T}&+...+(1)^n\alpha_0\alpha_n\mathbf{E}^0(\mathbf{E}^n)^{T}\\
&+...+\\
(1)^{n}\alpha_n\alpha_0\mathbf{E}^{n}(\mathbf{E}^{0})^{T}&+...+(1)^{2n}\alpha_n\alpha_n\mathbf{E}^{n}(\mathbf{E}^{n})^{T}
\end{aligned}
\right)
\end{aligned}
\end{equation}
\begin{equation}
\begin{aligned}
\mathbf{S}_{\mathbf{III}}
&=-\mathbf{E}_{\mathbf{III}}\mathbf{E}_{\mathbf{III}}^{T}\\
&=\left(
\begin{aligned}
(-1)^{3}\alpha_0\alpha_0\mathbf{E}^0(\mathbf{E}^0)^{T}&+...+(-1)^{n+3}\alpha_0\alpha_n\mathbf{E}^0(\mathbf{E}^n)^{T}\\
&+...+\\
(-1)^{n+3}\alpha_n\alpha_0\mathbf{E}^{n}(\mathbf{E}^{0})^{T}&+...+(-1)^{2n+3}\alpha_n\alpha_n\mathbf{E}^{n}(\mathbf{E}^{n})^{T}
\end{aligned}
\right)
\end{aligned}
\end{equation}
where $\mathbf{E}^{n}=\mathbf{\hat{A}}^{n}\mathbf{E}^{0}$ represents the node embedding of the $n$-th layer. We can extract the polynomials with odd-order subscript sums $\mathbf{S}_{ij}^{\mathbf{odd(I)}}=\alpha_i\alpha_j\mathbf{E}^{i}\mathbf{E}^{j}$, $\mathbf{S}_{ij}^{\mathbf{odd(III)}}=\alpha_i\alpha_j\mathbf{E}^{i}\mathbf{E}^{j}$ from $\mathbf{S}_{\mathbf{I}}$ and $\mathbf{S}_{\mathbf{III}}$. It is not difficult to see that them are equal $\mathbf{S}_{ij}^{\mathbf{odd(I)}}=\mathbf{S}_{ij}^{\mathbf{odd(III)}}$. We can extract the polynomials with even-order subscript sums $\mathbf{S}_{ij}^{\mathbf{even(I)}}=\alpha_i\alpha_j\mathbf{E}^{i}\mathbf{E}^{j}$, $\mathbf{S}_{ij}^{\mathbf{even(III)}}=-\alpha_i\alpha_j\mathbf{E}^{i}\mathbf{E}^{j}$ from $\mathbf{S}_{\mathbf{I}}$ and $\mathbf{S}_{\mathbf{III}}$, where $\mathbf{S}_{ij}^{\mathbf{even(I)}}=-\mathbf{S}_{ij}^{\mathbf{even(III)}}$. To prove that $\mathbf{S_{I}}=\mathbf{S_{III}}$ in the recommendation, we only need to prove that $\mathbf{S^{even(I)}}=\mathbf{S^{even(III)}}$ in the recommendation. 

\begin{figure}[ht]
\centering
\begin{minipage}{0.25\textwidth}
  \centering
  \includegraphics[width=1\linewidth]{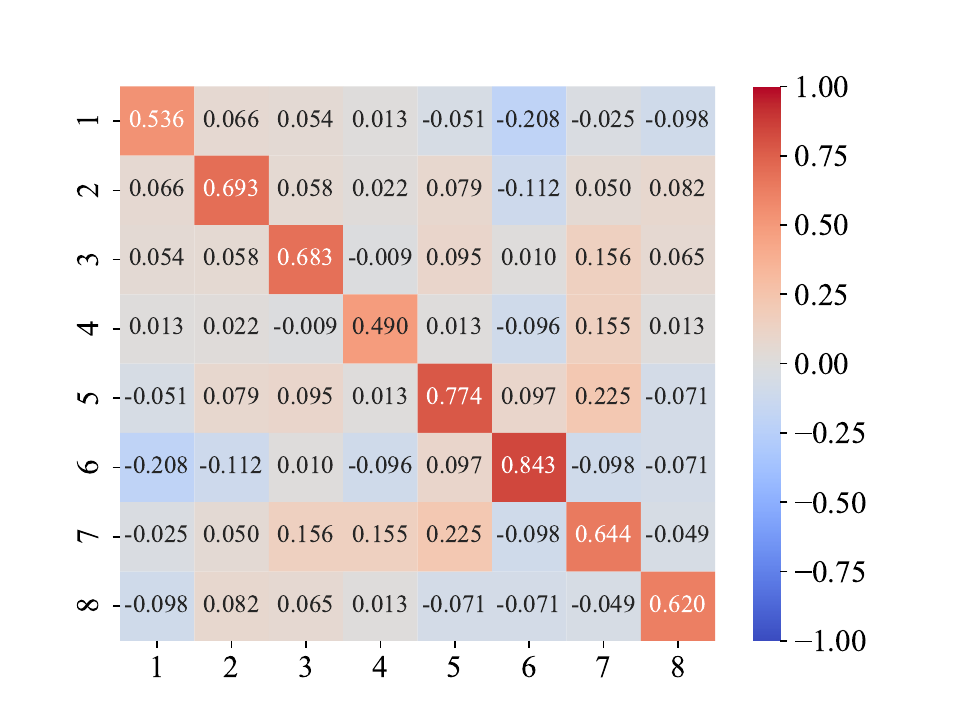}
  \centerline{(a) $\mathbf{E}^{0}\mathbf{E}^{0}$ in $\mathbf{S}^{\mathbf{even(I)}}$}
\end{minipage}%
\begin{minipage}{0.25\textwidth}
  \centering
  \includegraphics[width=1\linewidth]{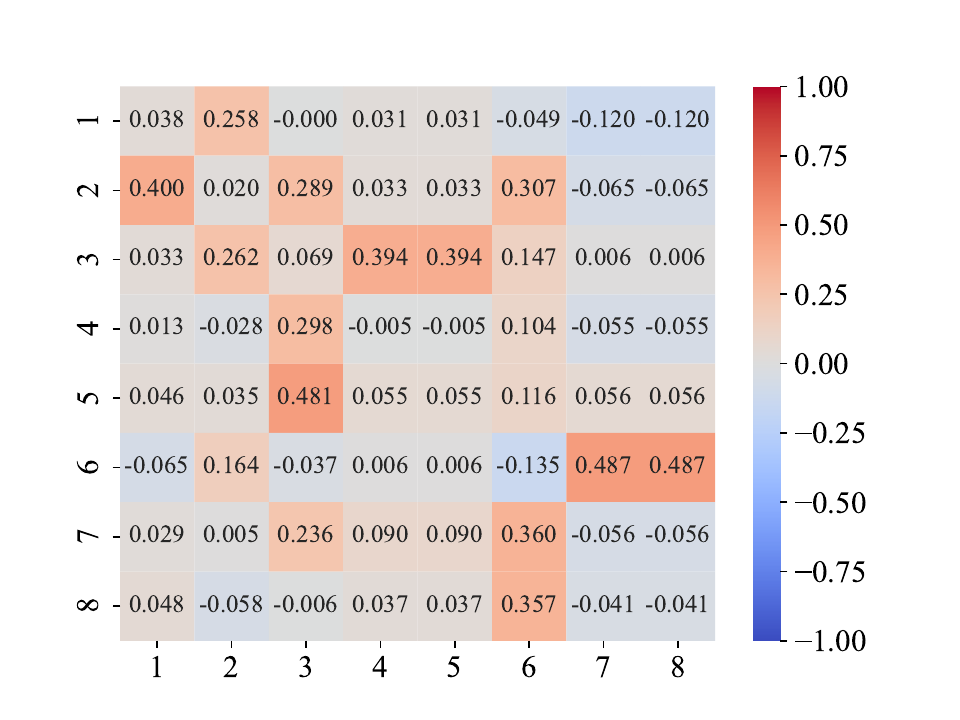}
  \centerline{(b) $\mathbf{E}^{0}\mathbf{E}^{1}$ in $\mathbf{S}^{\mathbf{odd(I)}}$}
\end{minipage}
\begin{minipage}{0.25\textwidth}
  \centering
  \includegraphics[width=1\linewidth]{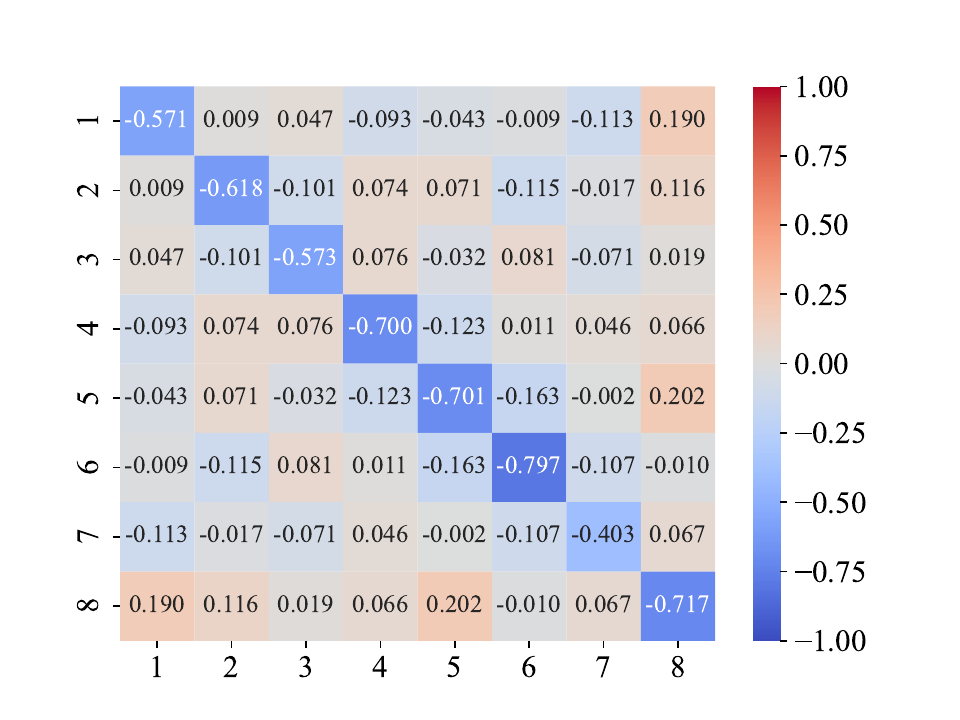}
  \centerline{(c) $\mathbf{E}^{0}\mathbf{E}^{0}$ in $\mathbf{S}^{\mathbf{even(III)}}$}
\end{minipage}%
\begin{minipage}{0.25\textwidth}
  \centering
  \includegraphics[width=1\linewidth]{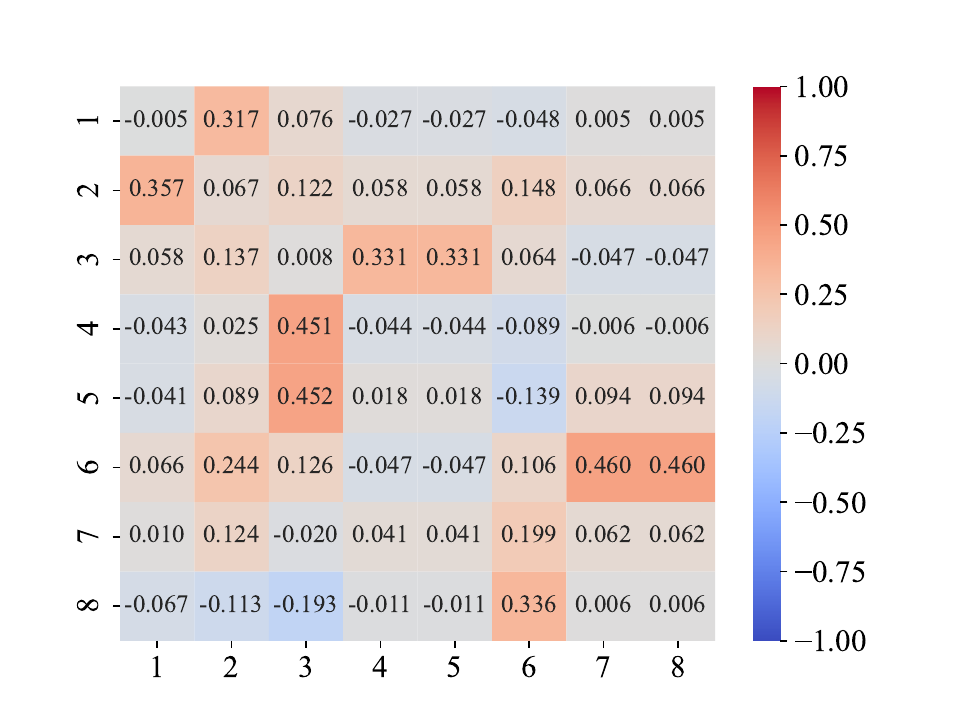}
  \centerline{(d) $\mathbf{E}^{0}\mathbf{E}^{1}$ in $\mathbf{S}^{\mathbf{odd(III)}}$}
\end{minipage}
\caption{The visualization results of $\mathbf{E}^{0}\mathbf{E}^{0}$ and $\mathbf{E}^{0}\mathbf{E}^{1}$ in graph embedding signals $\mathbf{S_{I}}$ and $\mathbf{S_{III}}$.}
\label{fig:pre}
\Description{ }
\end{figure}

In fact, we find that $\mathbf{S^{even(I)}}$ will increase the similarity between even-order neighbors on the graph, $\mathbf{S^{even(III)}}$ will reduce the similarity between even-order neighbors on the graph. We take $\mathbf{E}^{0}(\mathbf{E}^{0})^{T}$ and $\mathbf{E}^{0}(\mathbf{E}^{1})^{T}$ as examples, as shown in Figure \ref{fig:pre}. 
We find that 
$\mathbf{E}^{0}(\mathbf{E}^{0})^{T}$ in $\mathbf{S}^{\mathbf{even(I)}}$ emphasizes the similarity between the 0th-order node and its 0-order neighbors. For example, the similarity between node 1 and itself is 0.536. $\mathbf{E}^{0}(\mathbf{E}^{0})^{T}$ in $\mathbf{S}^{\mathbf{even(III)}}$ emphasizes the dissimilarity between the 0-order node and its 0-order neighbors. For example, the similarity between node 1 and itself is -0.576.

\begin{figure}[ht]
\centering
\begin{minipage}{0.25\textwidth}
  \centering
  \includegraphics[width=1\linewidth]{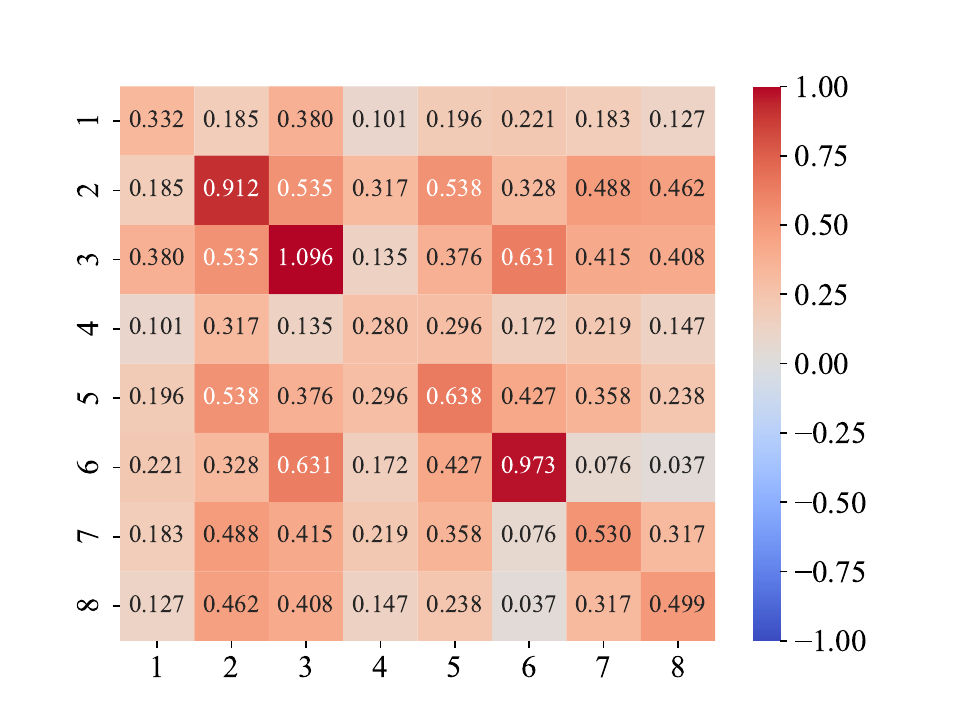}
  \centerline{(a) $\mathbf{E}^{0}\mathbf{E}^{0}$ in $\mathbf{S}^{\mathbf{even(I)}}$}
  \Description{ }
\end{minipage}%
\begin{minipage}{0.25\textwidth}
  \centering
  \includegraphics[width=1\linewidth]{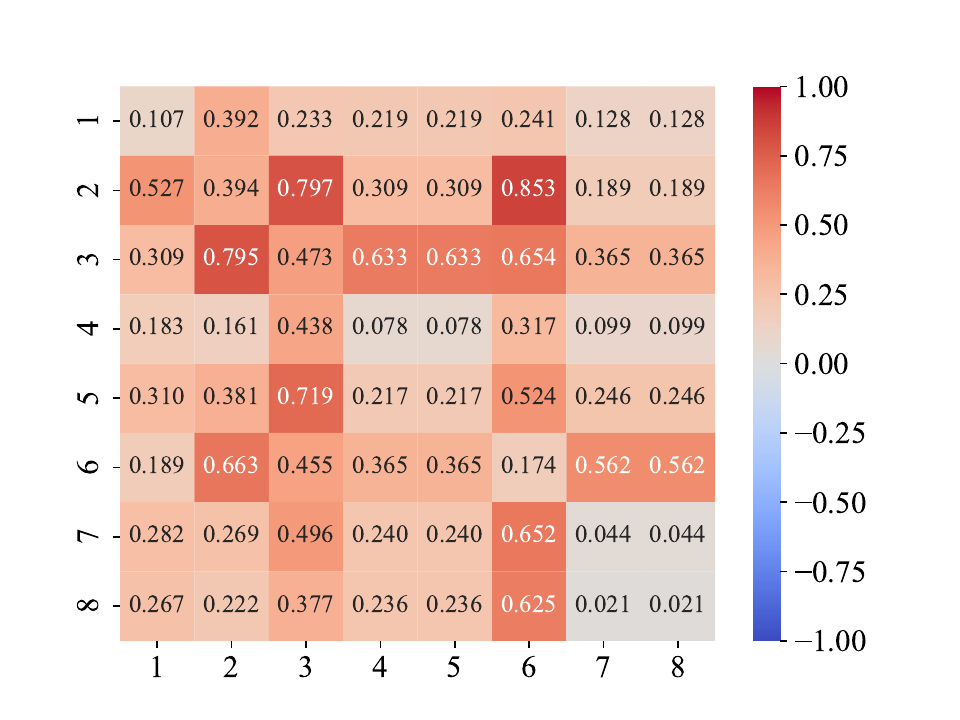}
  \centerline{(b) $\mathbf{E}^{0}\mathbf{E}^{1}$ in $\mathbf{S}^{\mathbf{odd(I)}}$}
  \Description{ }
\end{minipage}
\begin{minipage}{0.25\textwidth}
  \centering
  \includegraphics[width=1\linewidth]{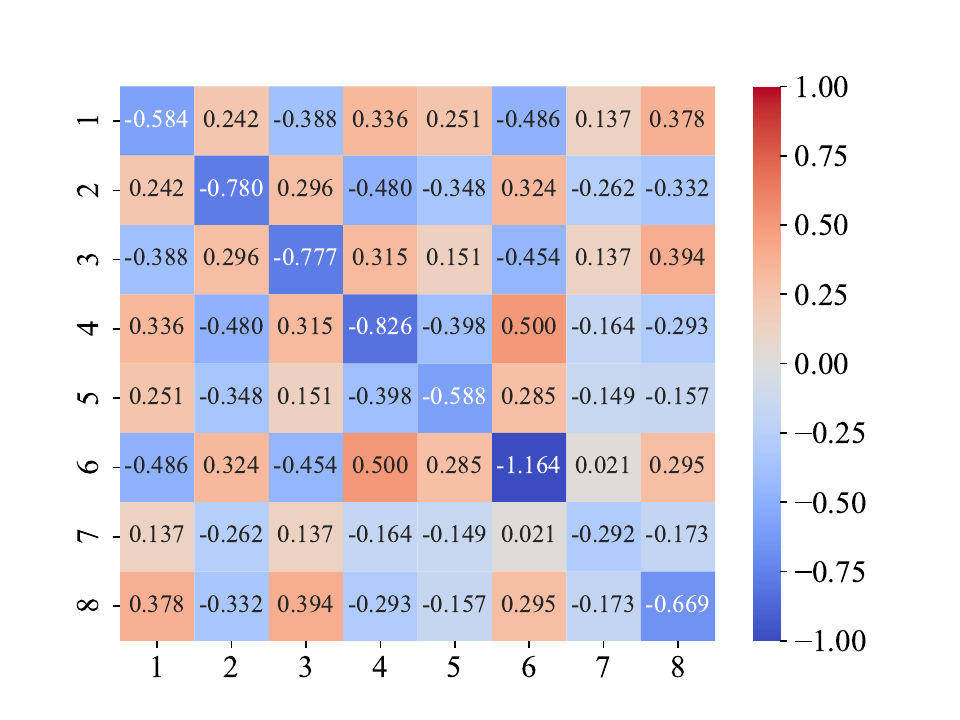}
  \centerline{(c) $\mathbf{E}^{0}\mathbf{E}^{0}$ in $\mathbf{S}^{\mathbf{even(III)}}$}
  \Description{ }
\end{minipage}%
\begin{minipage}{0.25\textwidth}
  \centering
  \includegraphics[width=1\linewidth]{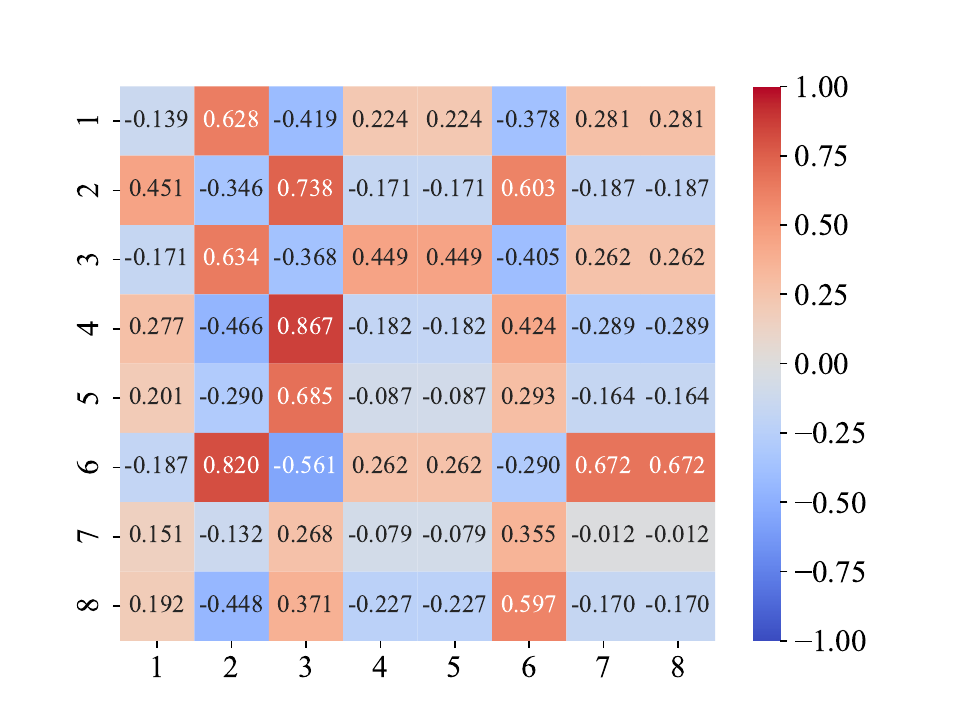}
  \centerline{(d) $\mathbf{E}^{0}\mathbf{E}^{1}$ in $\mathbf{S}^{\mathbf{odd(III)}}$}
  \Description{ }
\end{minipage}
\caption{The visualization results of training $\mathbf{E}^{0}\mathbf{E}^{0}$ and $\mathbf{E}^{0}\mathbf{E}^{1}$ in graph embedding signals $\mathbf{S_{I}}$ and $\mathbf{S_{III}}$ using BPR.}
\label{fig:aft}
\end{figure}

The loss function in recommendation emphasizes the similarity between users and items they have been exposed to, such as the BPR \cite{rendle2012bpr} loss:
\begin{equation}
\mathcal{L}_{bpr}=-\sum_{u\in\mathcal{U}}\sum_{i\in\mathcal{N}_u}\sum_{j\notin \mathcal{N}_u}\mathbf{ln}\sigma(\hat{y}_{ui}-\hat{y}_{uj})
\end{equation}
where $\hat{y}_{ui}$ represents the similarity between user $u$ and item $i$. $\mathcal{U}$ represents the set of users, and $\mathcal{N}_{u}$ represents the set of items that the user interacts with. BPR loss increases the similarity between interactive user-item pairs and suppresses the similarity between non-interactive user-item pairs, which is consistent with the purpose of recommendation. The $\mathbf{S^{even(I)}}$ and $\mathbf{S^{even(III)}}$ both emphasize the relationship between even-order neighbors, which does not conflict with the goal of recommendation. As shown in Figure \ref{fig:aft}, we use BPR loss to train $\mathbf{E}^0\mathbf{E}^0$ in $\mathbf{S^{even(I)}}$ and $\mathbf{S^{even(III)}}$. We can find that the original graph structure information and the information of BPR can be well preserved on $\mathbf{E}^0\mathbf{E}^0$, and they are orthogonal. During the training process of the recommendation model, them are equivalent, i.e. $\mathbf{S^{even(I)}}=\mathbf{S^{even(III)}}$ in recommendation. Therefore, the low-frequency grapn signal $\mathbf{S_{I}}$ and the high-frequency grapn signal $\mathbf{S_{III}}$ with the same absolute value have equal effects on the recommendation.

\end{document}